%% file: main.tex
\pdfoutput=1

\documentclass[11pt]{article}

\usepackage[final]{acl}

\usepackage{times}
\usepackage{latexsym}
\usepackage{caption}
\usepackage[T1]{fontenc}

\usepackage[utf8]{inputenc}

\usepackage{microtype}

\usepackage{inconsolata}

\usepackage{graphicx}

\newcommand{\ignore}[1]{}

\usepackage{subfiles}
\usepackage{xcolor}
\usepackage{booktabs, multirow}
\usepackage{soul}
\usepackage{makecell}
\usepackage{tcolorbox} 
\newtcolorbox{fullwidthbox}{
    colback=gray!10, 
    colframe=black, 
    width=\textwidth, 
    sharp corners, 
    boxrule=1pt, 
    left=5pt, right=5pt, top=5pt, bottom=5pt 
}
\usepackage{titletoc}
\usepackage{todonotes}
\newcommand{\SideNote}[2]{\todo[color=#1,size=\small]{#2}}
\newcommand{\violet}[1]{\SideNote{purple!40}{#1 --Violet}}

\title{FLAMES: Improving LLM Math Reasoning \\ via a Fine-Grained Analysis of the Data Synthesis Pipeline}

\author{\textbf{Parker Seegmiller}$^{\ddagger}$\thanks{Work done while interning at Amazon AGI Foundations.}, \textbf{Kartik Mehta}$^{\dagger}$, \textbf{Soumya Saha}$^{\dagger}$, \textbf{Chenyang Tao}$^{\dagger}$, \\
\textbf{Shereen Oraby}$^{\dagger}$, \textbf{Arpit Gupta}$^{\dagger}$, \textbf{Tagyoung Chung}$^{\dagger}$, \\
\textbf{Mohit Bansal}$^{\dagger\S}$, \textbf{Nanyun Peng}$^{\dagger\P}$ \\
$^{\dagger}$Amazon AGI Foundations, $^{\ddagger}$Dartmouth College, \\
$^{\S}$UNC Chapel Hill, $^{\P}$University of California, Los Angeles \\
\texttt{pkseeg.gr@dartmouth.edu} \quad \texttt{kartim@amazon.com}
}

\begin{document}
\maketitle

\subfile{sections/00_Abstract.tex}

\subfile{sections/01_Introductionv2.tex}
\subfile{sections/03_Framework.tex}
\subfile{sections/04_Agents_Short.tex}

\subfile{sections/05_QualityControl.tex}
\subfile{sections/07_Results.tex}

\subfile{sections/08_Conclusion.tex}
\subfile{sections/09_Limitations.tex}
\bibliography{custom}

\newpage
\appendix

\subfile{sections/AA_Appendixv2.tex}

\end{document}

%% file: sections/00_Abstract.tex
\begin{abstract}
Recent works improving LLM math reasoning with synthetic data have used unique setups, making comparison of data synthesis strategies impractical. This leaves many unanswered questions about the roles of different factors in the synthetic data pipeline, such as the impact of filtering low-quality problems. To address this gap, we introduce FLAMES, a \underline{F}ramework for \underline{L}LM \underline{A}ssessment of \underline{M}ath r\underline{E}asoning Data \underline{S}ynthesis, and perform a systematic study of $10$ existing data synthesis strategies and multiple other factors impacting the performance of synthetic math reasoning data. Our FLAMES experiments provide several valuable insights about the optimal balance of difficulty and diversity of synthetic data. \underline{First}, data agents designed to increase problem complexity lead to best improvements on most math metrics. \underline{Second}, with a fixed data generation budget, keeping higher problem coverage is more important than keeping only problems with reliable solutions. \underline{Third}, GSM8K- and MATH-based synthetic data can lead to improvements on competition-level benchmarks, showcasing easy-to-hard generalization. Leveraging insights from our FLAMES experiments, we design two novel data synthesis strategies for improving out-of-domain generalization and robustness. Further, we develop the FLAMES dataset, an effective blend of our novel and existing data synthesis strategies, outperforming public datasets on OlympiadBench (+15.7), CollegeMath (+4.5), GSMPlus (+6.5), and MATH (+3.1). Fine-tuning Qwen2.5-Math-7B on the FLAMES dataset achieves 81.4\% on MATH, surpassing larger Llama3 405B, GPT-4o and Claude 3.5 Sonnet.
\end{abstract}

%% file: sections/01_Introductionv2.tex
\begin{figure}[t]
  \includegraphics[width=\columnwidth]{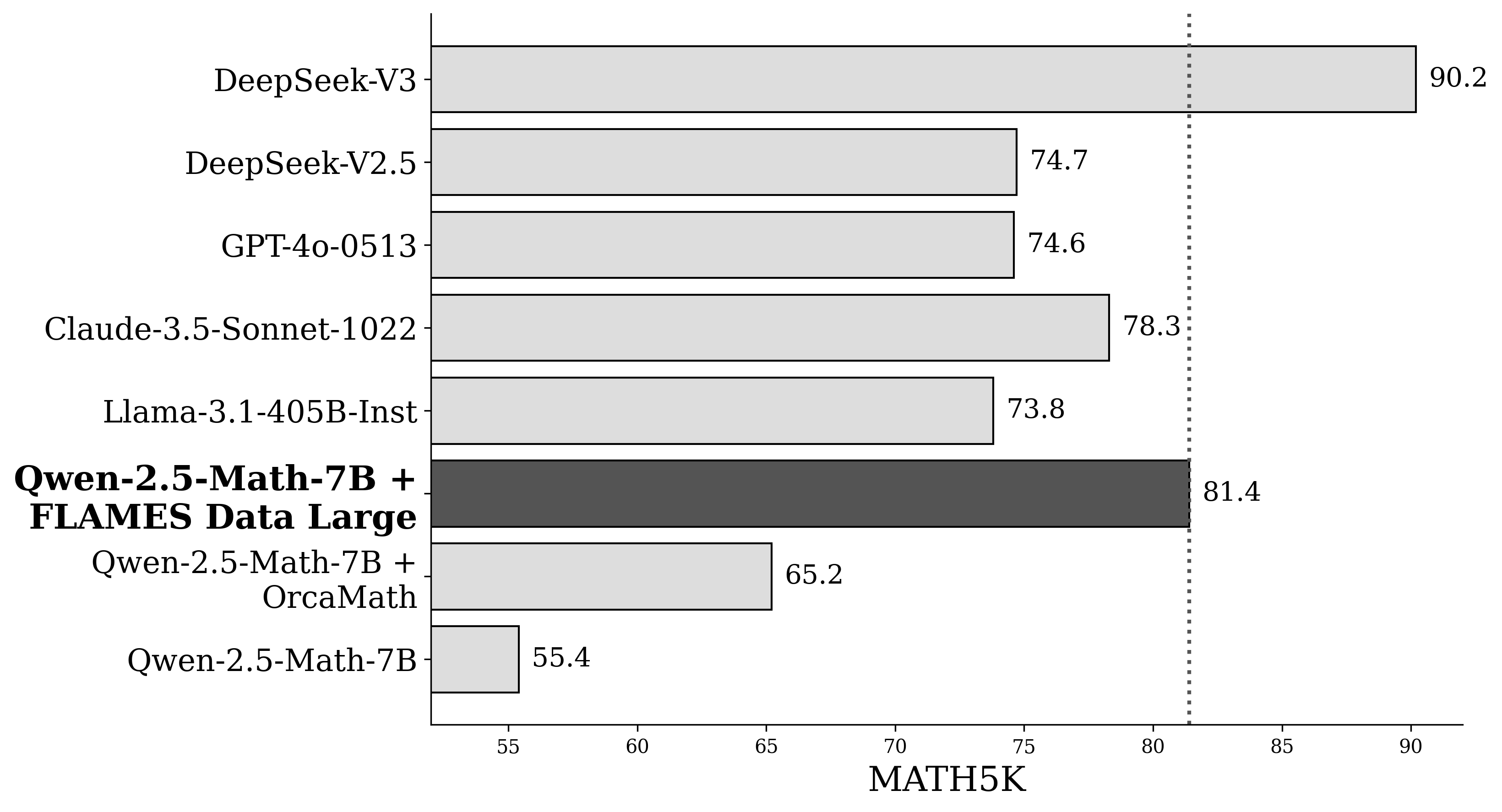}
  \caption{MATH benchmark scores for popular LLMs. Qwen2.5-Math-7B + X (FLAMES or OrcaMath) denotes results obtained by finetuning Qwen2.5-Math-7B model with X dataset. Comparison of FLAMES data with other public Math datasets is shown in Tables \ref{tab:open-source}, \ref{tab:underlying}.}
  \label{fig:experiments}
\end{figure}

\section{Introduction}
\begin{figure}[t]
  \includegraphics[width=\columnwidth]{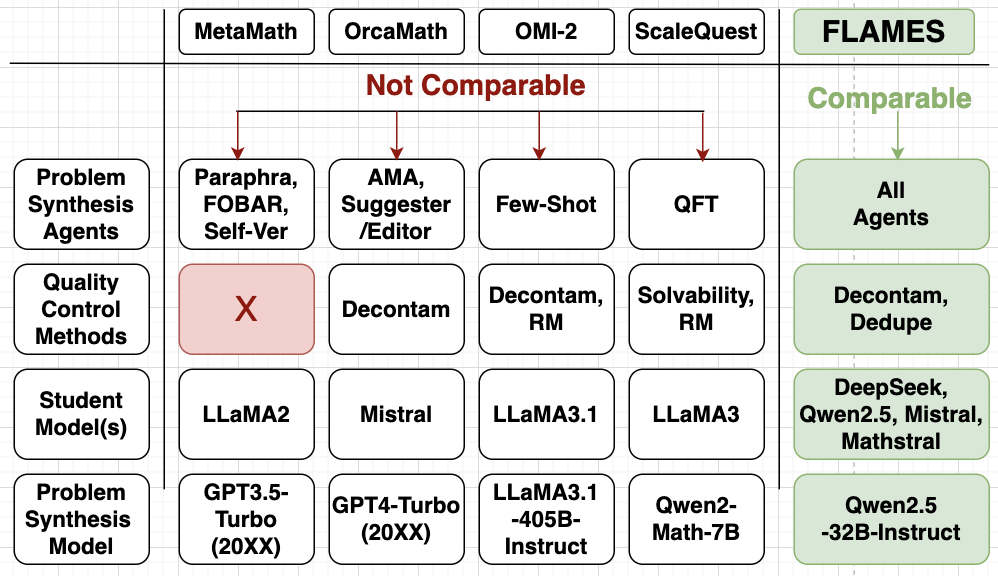}
  \caption{Landscape of recent math data synthesis works. Each work uses non standardized setups, such as different synthesis models, student models, and quality control, making comparison across works impractical. Detailed version is shown in Appendix Figure \ref{fig:comparison_appendix}}
  \label{fig:preview}
\end{figure}

Solving math problems is a key measure of evaluating the reasoning ability of large language models (LLMs) \ignore{in the field of natural language processing (NLP)} \cite{wei2022chain, cobbe2021training, hendrycks2measuring}. Due to the challenges of creating a large-scale human-crafted dataset, LLM-based generation of synthetic data has been explored and proven effective in improving LLM math reasoning capabilities \cite{metamath, kpdds, scalequest}. While early works on math reasoning data synthesis focused on synthesis of solutions (reasoning traces)\ignore{\violet{does solutions refer to the final answer or the reasoning paths? We should distinguish the two.}} for existing problems, e.g. rejection sampling \cite{Tong2024DARTMathDR}, recent works have shifted focus towards increasing problem difficulty and diversity by synthesizing new problems. Recent works such as Orca-Math \cite{orcamath}, MetaMath \cite{metamath}, and OpenMath-Instruct-2 \cite{omi2} have proposed one or more strategies\footnote{Data synthesis strategies often use one or multiple LLM calls and are commonly referred to as data agents. We use the terms \textit{strategies} and \textit{agents} interchangeably in this paper.} for generating new math problems. However, these works often use different setups (see Figure \ref{fig:preview}),\ignore{\violet{can you give some examples of the "setup" you refer to?}} making the comparison of data synthesis strategies across studies infeasible. This lack of standardization deprives researchers and practitioners of practical insights, leaving them uninformed about the real factors driving performance improvements via math reasoning data synthesis. 


Figure \ref{fig:preview} summarizes setups of four popular math reasoning data synthesis works in terms of four experimental factors, highlighting the lack of standardization. As different problem synthesis models and student models have been used in the OrcaMath and ScaleQuest papers, for example, it is infeasible to conclude whether Suggester-Editor \cite{orcamath} is a better data synthesis strategy or question fine-tuning (QFT) \cite{scalequest}. This lack of standardization poses several important research questions about math reasoning data synthesis, which are currently under-studied. \underline{\textbf{RQ1:}} How do different strategies of data quality control affect the performance of synthetic data, and what strategy works the best? \underline{\textbf{RQ2:}} Which data synthesis strategy performs optimally for improving student model math reasoning? \underline{\textbf{RQ3:}} With a fixed compute budget for synthetic problem generation, should we mix data from multiple strategies or just scale the best strategy? \underline{\textbf{RQ4:}} How does the choice of problem and solution generation models impact performance with math synthetic data?

\begin{figure*}[t]
 \centering
  \includegraphics[width=1.8\columnwidth]{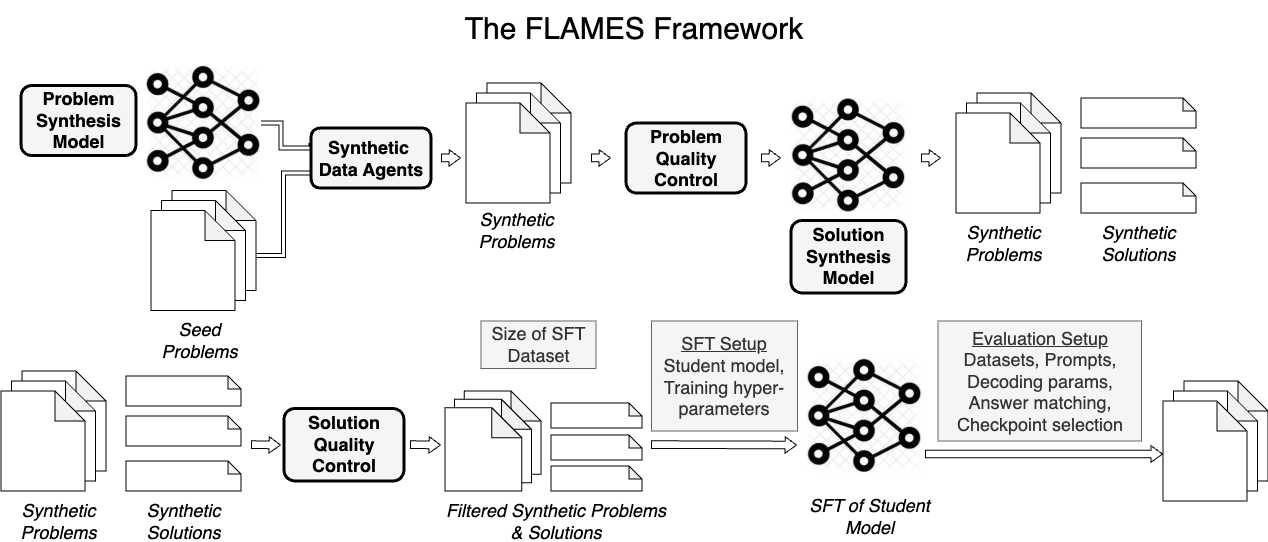}
  \caption{Overview of the FLAMES framework, showing fine-grained components of the Math data synthesis pipeline. Components for which we run \textbf{controlled experiments} are shown in bold font.}
  \label{fig:overview}
\end{figure*}

To \ignore{address the lack of a unified experimental setup and}study these open research questions, we propose the \underline{F}ramework for \underline{L}LM \underline{A}ssessment of \underline{M}ath r\underline{E}asoning with Data \underline{S}ynthesis (FLAMES). Figure \ref{fig:overview} shows a flow diagram of the FLAMES framework. In this framework, we create a comprehensive list of the factors that may impact performance within the synthetic data pipeline, and we choose fixed values for them based on existing literature and our experimental findings. The FLAMES framework enables controlled experiments where we vary only one factor, keeping other factors fixed. We perform a controlled study of ten existing data synthesis strategies, six data quality control strategies, two problem generation models, and two solution generation models. These experiments provide many valuable insights into data synthesis for LLM math reasoning\ignore{, including the need to balance difficulty and diversity of the synthetic data}. \underline{\textbf{First}}, we find that data agents designed to increase problem complexity lead to best improvements on most math metrics. \underline{\textbf{Second}}, with a fixed data generation budget, keeping higher coverage of synthetic problems (even with some inaccuracy) is more important than keeping only problems with reliable solutions. \underline{\textbf{Third}}, synthetic data generated based on GSM8K and MATH can lead to improvements on competition-level benchmarks, showcasing easy-to-hard generalization capability. \underline{\textbf{Fourth}}, choice of solution generation model impacts student model performance more than choice of problem generation model.

Leveraging the findings from our study of existing data synthesis strategies (Table \ref{tab:agents}), we identify out-of-domain (OOD) reasoning \footnote{All ten agents lead to CollegeMath scores less than 41.0.} and robustness to distracting information \footnote{At least 13 points difference between performance on GSM8K and GSMplus-Distraction for all ten agents.} as two weaknesses in existing data synthesis strategies. We design novel agents, \textbf{Taxonomy-Based Key Concepts} and \textbf{Distraction Insertion}, for addressing these weaknesses. We demonstrate with the FLAMES framework that the data generated using these two agents leads to superior performance gains on relevant OOD and distraction benchmarks. \ignore{CollegeMath (evaluating out-of-domain) and GSMPlus distraction (evaluating robustness), respectively.}

Further, we study the impact of mixing data from different agents, and design an effective blend of agent data. We show empirically that this blend (FLAMES Small) leads to balanced performance across five evaluation datasets, with performance surpassing all ten existing agents on OOD and competition-level benchmarks. We scale this blend to construct the FLAMES Large (1M) and FLAMES XL (1.5M) datasets. Our experiments across five student models (Table \ref{tab:underlying}), show that FLAMES Large leads to better performance than any existing public math dataset. With FLAMES XL, we observe drastic improvements over all public datasets across OlympiadBench (+12.8), GSMplus (+4.9), CollegeMath (+5.8) and Math (+1.3). Fine-tuning Qwen2.5-Math-Base on the FLAMES Large dataset leads to 81.4\% Math, surpassing larger Llama3 405B, GPT-4o and Claude 3.5 Sonnet (see Figure \ref{fig:experiments}). These performance gains can be attributed to 1) findings from our FLAMES framework experiments on problem quality control, data agents, and teacher models, 2) our two novel data agents leading to improved performance on OOD and robustness metrics, and 3) our effective blend of data agents leading to balanced performance across all evaluation datasets.

In summary, our paper makes four contributions.\\
\underline{\textbf{1. }} We propose the FLAMES framework for controlled study of multiple factors involved in the math synthetic data pipeline. This enables systematic comparison of existing and future math data agents.\\
\underline{\textbf{2. }} We leverage the FLAMES framework to perform analysis of $12$ data agents, $6$ data quality control strategies, $2$ problem generation and $2$ solution generation models. Our experiments provide valuable insights for researchers working on improving LLM math reasoning with synthetic data. \\
\underline{\textbf{3. }} We design $2$ novel data synthesis agents, and empirically establish that they lead to enhanced robustness and out-of-domain improvements compared to existing data agents.\\
\underline{\textbf{4. }} We develop the FLAMES dataset, consisting of an effective blend of our $2$ novel agents and $2$ strong existing data agents, showing drastic improvements over existing public math datasets.





%% file: sections/03_Framework.tex
\section{The FLAMES Framework}
\label{sec:framework}
Figure \ref{fig:overview} shows a flow diagram of the FLAMES framework. There are multiple factors that play key roles in final student model performance after fine-tuning with synthetic data. As part of the FLAMES framework, we list these factors and provide their standard values based on either controlled experiments or existing literature. We highlight the factors for which we perform controlled experiments in Figure \ref{fig:overview}, and discuss those briefly in this section. We refer readers to Appendix \ref{app:framework-details} for a detailed discussion of these factors.

\textbf{Problem Synthesis Agents:} Math data synthesis agents interact with one or more seed problems and a problem generation model to synthesize novel math problems. These data agents play a crucial role in the synthetic math data pipeline, and are the focus of many recent works \cite{scalequest, metamath, orcamath}. In our work, we use the train splits from the popular GSM8K \cite{cobbe2021training} and MATH \cite{hendrycks2measuring} data for seed problems. We discuss details of 10 existing and 2 novel data agents in Section \ref{sec:data-agents}.

\textbf{Quality Control:} All problems generated in the FLAMES framework are first deduplicated using exact match. We remove any synthetic problem if there is a high amount of token overlap between the synthetic problem and a test set problem. We study additional data quality control techniques in Section \ref{sec:quality-control}.

\textbf{Student Model:} Choice of student model plays a crucial role for studying LLM math reasoning improvements with synthetic data. DeepSeek-Math-7B \cite{shao2024deepseekmath} is a relatively-performant non-instruction-tuned base model, and is commonly used for measuring LLM math reasoning with synthetic data \cite{scalequest, kpdds}. We use this model as the underlying student model for the FLAMES framework.

\textbf{Problem Synthesis Model:} To generate problems, data agents interact with Qwen2.5-32B-Instruct \cite{yang2024qwen2}, which is chosen based on a controlled experiment (see Section \ref{sec:synth-model-comparison}).


\textbf{Solution Synthesis:} An LLM acts as a solution generation teacher model providing solutions for synthetic problems. Qwen2.5-Math-7B-Instruct \cite{yang2024qwen2} has been shown to offer a good blend of performance and speed, scoring 95.2 and 83.6 on GSM8K and MATH, respectively. We use Qwen2.5-Math-7B-Instruct as the solution synthesis model in the FLAMES framework. We compare alternative models in Section \ref{sec:synth-model-comparison}.

\textbf{Evaluation of Student Models:} We evaluate fine-tuned student models on 5 datasets across 4 categories. For \underline{\textit{in-domain evaluation}}, we use the GSM8K (1,319 samples) \cite{cobbe2021training} and the MATH5K test sets (5,000) \cite{hendrycks2measuring}. These datasets are considered in-domain as their training datasets are used as seed problems for synthetic problem generation. To evaluate fine-tuned models on an \underline{\textit{out-of-domain}} set of problems which are similar in grade-level to the seed problems (MATH and GSM8K), we use the CollegeMath test set (2,818) \footnote{\url{https://huggingface.co/datasets/qq8933/College_Math_Test}}, which contains exercises sourced from university-level textbooks across several math subjects. For \underline{\textit{robustness}} evaluation, we use GSMPlus dataset \cite{li2024gsm}, an adversarial grade school math dataset which includes GSM8K questions subject to numerical variation, arithmetic variation, and paraphrasing. We exclude the critical thinking portion for easy evaluation and separately report metrics for the distraction insertion subset. To evaluate \underline{\textit{competition-level}} reasoning, we use OlympiadBench \footnote{\url{https://huggingface.co/datasets/realtreetune/olympiadbench}} (675), which contains competition math problems covering difficult topics such as combinatorics and number theory. 

\textbf{Training Details:} \ignore{We use the DeepSeek-Math-7B model \cite{shao2024deepseekmath} as the student model in the FLAMES framework.} We do full-parameter fine-tuning on Amazon EC2 P4 instances\footnote{\url{https://aws.amazon.com/ec2/instance-types/p4/}} using Deepspeed Zero3 \cite{rajbhandari2020zeromemoryoptimizationstraining} for distributed training across 8 A100 GPUs. We train using a batch size of 4 for 5 epochs, saving 10 checkpoints. We implement SFT using the SWIFT library\ignore{\footnote{\url{https://github.com/modelscope/ms-swift}}} \cite{zhao2024swiftascalablelightweightinfrastructure}, using default hyperparameters.

%% file: sections/04_Agents_Short.tex
\section{Data Synthesis Agents}
\label{sec:data-agents}
We select $10$ existing problem generation agents based on recent works in math data synthesis. \ignore{(refer to Appendix \ref{app:agent-details} for details)} Based on the primary objective of these agents, we categorize them as belonging to one of four categories: 1) In-domain practice, 2) In-domain complexity enhancing, 3) Robustness enhancing, and 4) Out-of-domain enhancing. Additionally, based on experimental findings detailed in Section \ref{sec:data-agents-exp}, we propose two novel agents under the categories of robustness enhancing and out-of-domain enhancing. We briefly describe these 12 agents here, and refer readers to Appendix \ref{app:agent-details}  for details. 

\subsection{Existing Agents}
\underline{\textit{In-Domain Practice}} agents are designed to increase the quantity of in-domain data available for model training. We implement the \textbf{Few-Shot} agent \cite{omi2} and the \textbf{Paraphrasing} agent \cite{metamath}, along with the \textbf{Key Concepts} and \textbf{Seeded Key Concepts} agents \cite{kpdds}.

\underline{\textit{In-Domain Complexity Enhancing}} agents aim to increase complexity of existing math word problems \cite{orcamath, mmiqc, luo2023wizardmath}. We implement the \textbf{Suggester-Editor} agent \cite{orcamath} and the \textbf{Iterative Question Composing (IQC)} agent \cite{mmiqc}. \ignore{for this category}

\underline{\textit{Robustness Enhancing}} agents aim to alter existing problems to increase the robustness of the student model's reasoning process. These agents include the \textbf{Ask Me Anything} agent \cite{orcamath}, the \textbf{Self-Verification} agent \cite{metamath}, and the \textbf{FOBAR} agent \cite{metamath}.

\underline{\textit{Out-of-Domain Enhancing}} agents are not directly seeded with in-domain problems. We implement the \textbf{Question Fine-Tuning (QFT)} agent \cite{scalequest} for this category.

\subsection{Novel Agents}
We propose the novel \textbf{Taxonomy-Based Key Concepts} and \textbf{Distraction Insertion} agents, targeting out-of-domain generalization and robustness to distraction, respectively.  

The \underline{\textbf{Taxonomy-Based Key Concepts}} agent \ignore{follows \cite{kpdds, nvidia2024nemotron4340btechnicalreport} by performing} performs a two step generation of synthetic problems based on a novel curated taxonomy of math subjects. The first step generates a list of key concepts relevant to a given math subject, followed by the second step of generating novel problems based on each key concept. We curate our math taxonomy by combining taxonomies from a variety of sources \cite{kpdds, liu2024mathbenchevaluatingtheoryapplication, huang2024mustardmasteringuniformsynthesis, didolkar2024metacognitivecapabilitiesllmsexploration}. This agent can be considered an out-of-domain enhancing agent, as it uses no seed problems, unlike other key concepts agents \cite{kpdds}.


The \underline{\textbf{Distraction Insertion}} agent inserts distracting information into an existing problem, without changing problem-relevant information. This agent is designed to improve reasoning performance in the presence of adversarial distracting information, which needs to be ignored for solving the problem. 

See Figure \ref{fig:agents} for examples and Appendix \ref{sec:promptlocation} for the location of relevant prompts.


%% file: sections/05_QualityControl.tex
\section{Data Quality Control}
\label{sec:quality-control}

A challenge for LLM-based math problem synthesis is in controlling the quality of synthetic data, i.e. the removal of unsolvable or ill-formatted problems and incorrect solutions. Existing works for math data synthesis primarily rely on proprietary models like GPT-4 for problem and solution generation and assume that generated data is of good quality. Consequently, the impact of data quality control for math synthetic data has been an under-studied problem. Our work is the first to study this factor in a principled way and provide insights on what works best for math synthetic data. 

Recent work \cite{scalequest} has introduced the use of an LLM-based solvability filter to remove ill-formatted or unsolvable problems and  a reward model (InternLM2-7B-Reward) to select the preferred solution. Self-consistency \cite{wang2023h} is another viable alternative for ensuring problem and answer quality, where multiple (three in our case) solutions are generated for a problem with temperature sampling and only those problems are kept where at least a pre-determined number of the solutions lead to the same answer.

We use these to design six data filtering strategies for our experiments. In \underline{\textbf{Strict Self-Consistency}}, only problems with a matching answer in all $3$ solutions are kept and one solution is randomly selected. \underline{\textbf{Majority Self-Consistency}} (or Majority) keeps problems where at least $2$ solutions contain matching answers, and one of those solutions is randomly selected. \underline{\textbf{Solvability + RM}} first filters problems which are deemed ``unsolvable\footnote{The solvability filtering prompt can be found in Appendix \ref{app:agent-prompts}.}'' by the Qwen2.5-Math-7B-Instruct model, then selecting the solution which receives the highest reward according to the InternLM2-7b-Reward reward model (RM) \cite{cai2024internlm2}. \underline{\textbf{Majority + First}} is similar to Majority Self-Consistency, but uses the first solution in case no majority is found.  This method doesn't remove any problems as part of the data quality control. \underline{\textbf{Solvability + First}} filters unsolvable problems, then includes the first solution for each problem considered solvable. \underline{\textbf{First}} keeps the first generated solution for each synthetic problem, keeping all the problems. 

We note that these different techniques are designed to remove noisy synthetic data, but may also end up removing some legitimate problems and solutions, especially harder problems. We choose these six strategies to evaluate a diverse array of coverage, difficulty and accuracy levels in filtered synthetic datasets. We discuss findings of our systematic study of these strategies in Section \ref{sec:quality-control-exp}.

%% file: sections/07_Results.tex
\section{Experimental Results}
\subfile{../tables/Quality_Controlv2.tex}

Here we explore quality control methods for synthetic math reasoning data (Section \ref{sec:quality-control-exp}). We utilize the FLAMES framework to compare math reasoning data synthesis strategies (Section \ref{sec:data-agents-exp}). We introduce and show the effectiveness of the FLAMES datasets compared with existing math datasets (Section \ref{sec:os-dataset-exp} and Section \ref{sec:underlying}). Last, we evaluate the relative impact of problem and solution generation teacher models (Section \ref{sec:synth-model-comparison}).

\subsection{Quality Control of Synthetic Data}
\label{sec:quality-control-exp}
To study the impact of quality control strategies in the math synthetic data pipeline, we perform fine-tuning of the DeepSeek-Math-7B model using datasets with varying types of quality control applied. We start with 150k synthetic problems generated using the Suggester-Editor agent\footnote{We use Suggester-Editor data agent for this study, as we find this agent to be highly performant.} with the Qwen2.5-32B-Instruct model, and synthesize three independent solutions to each problem using the Qwen2.5-Math-7B-Instruct model. After applying each quality control method from Section \ref{sec:quality-control} for removing problems and selecting solutions, we use the resulting dataset for finetuning. Finally, we evaluate the fine-tuned model on in-domain test sets. We perform two experiments: \underline{\textbf{varying coverage}}, where we use the resulting dataset after filtering, and \underline{\textbf{fixed coverage}}, where we subsample each filtered dataset to have the same training size (45K)\footnote{45K matches the size of the smallest dataset (Strict Self-Consistency) after filtering.}.

\subfile{../tables/Agents.tex}

Table \ref{tab:quality-control} shows results of this experiment.  We observe four key findings. \textbf{\underline{First}}, we find that solvability filtering (in its current form) is not effective as \textit{Solvability + First} (R3) shows inferior performance to \textit{First} (R5). This is likely due to filtering out genuine problems with the solvability filter. Using the solvability filter on the human-crafted MATH test set, we observe that it removes 30\% of the real problems (30\% and 50\% from difficulty levels 4 and 5, respectively), indicating that this filtering is unreliable (see Appendix \ref{app:solvability} for details). \textbf{\underline{Second}}, coverage of problems matters more than the accuracy of the solutions. We observe superior performance with \textit{Majority + First} (R6) than \textit{Majority} (R4), where the only difference (between R6 and R4) is inclusion of additional 60K problems in R6 where first solution is used. This shows that including more problems, even with potentially incorrect solutions, leads to better performance. \textbf{\underline{Third}}, with the same scale and problem complexity, precision of the solution matters, as \textit{Majority + First} (R6) shows better performance than \textit{First} (R5) strategy. This is also evident from \textit{fixed coverage} analysis, where we observe higher Math scores for strategies designed for higher precision\footnote{However, we observe less variation and a different trend on GSM8K problems. This is likely due to smaller difference between solutions with First and Majority strategy on relatively easier GSM8K level problems.}. \textbf{\underline{Fourth}}, \textit{Majority + First} (R6) provides a good blend of coverage of problems and accuracy of the solutions, and superior performance than other strategies. With comparable performance, the \textit{First} (R5) strategy is computationally better than the \textit{Majority + First} (R6). Hence, we use \textit{First} strategy for all our further studies.



\subsection{Comparison of Data Generation Agents}
\label{sec:data-agents-exp}
To assess which synthetic data agents generate the most performant data, we perform SFT of the DeepSeek-Math-7B base model with datasets generated using each agent from Section~\ref{sec:data-agents}. We generate 150K unique problems with each agent, using the \textit{First} strategy for data quality control. We compare data agents by evaluating performance of their respective fine-tuned DeepSeek-Math-7B models. 

Table \ref{tab:agents} shows the results of each fine-tuned model across the $5$ benchmark datasets. We observe four major findings from this study. \textbf{\underline{First}}, we observe that complexity-enhancing agents (IQC and Suggester-Editor) lead to best improvement on in-domain metrics. Surprisingly, they also lead to best competition-level and robustness improvements. \textbf{\underline{Second}}, the out-of-domain enhancing agents (particularly, our novel Taxonomy Key Concepts agent) lead to best improvement on out-of-domain reasoning, surpassing complexity-enhancing agents. \textbf{\underline{Third}}, we find that our novel Distraction Insertion agent leads to best performance on the distraction insertion benchmark within the GSMPlus dataset, showcasing that the FLAMES framework enables design of agents to solve specific reasoning tasks. \textbf{\underline{Fourth}}, we observe that data agents, while using only GSM8K and MATH as seed sets, lead to improvement over competition level benchmarks (as evident from OlympiadBench metrics). This is particularly interesting as it demonstrates "easy-to-hard generalization", and provides evidence that math data synthesis may be used to improve performance on more difficult math reasoning tasks.

\subsection{Comparison of FLAMES Datasets}
\label{sec:os-dataset-exp}
We study the impact of mixing data from different agents (refer to experiments in Appendix \ref{app:combining-agents}) and observe best results by mixing 50\% Suggester-Editor, 20\% IQC, 20\% Taxonomy Key Concept (our proposed) and 10\% Distraction Insertion (our proposed) agents. Based on this blend, we develop 3 versions of FLAMES datasets --- FLAMES Small (150K), FLAMES Large (1M), and FLAMES XL (1.5M). We compare FLAMES Small with individual agents in Table \ref{tab:agents}, and observed balanced performance across the five evaluation datasets, with performance surpassing all $12$ agents on CollegeMath and Olympiad-Bench.
\subfile{../tables/OpenSourceComp_Short.tex}

For comparing the performance of FLAMES Large and FLAMES XL datasets, we conduct finetuning of the DeepSeek-Math-7B model on a diverse set of existing math datasets, including GSM8K \cite{cobbe2021training}, MATH \cite{hendrycksmeasuring}, NuminaMath \cite{numina_math_datasets}, MetaMathQA \cite{metamath}, OrcaMath \cite{orcamath}, OpenMathInstruct2\footnote{We include only unique problems from OpenMathInstruct2 for fair comparison, since ScaleQuest and FLAMES datasets contain unique problems.} \cite{omi2}, MMIQC \cite{mmiqc}, and ScaleQuest \cite{scalequest}. Since different synthetic datasets were synthesized with different solution generation models, we compare results with "refreshed" versions of each synthetic dataset, where one solution for each unique problem is generated using Qwen2.5-Math-7B-Instruct (same as used for our FLAMES dataset, for fair comparison). Results with original datasets, which performed worse in each case, are given in Appendix \ref{app:open-source}. 


Table \ref{tab:open-source} shows results for this study. We observe that ScaleQuest (refreshed with Qwen2.5-Math-7B-Instruct solutions) leads to best average performance (60.0) among the public datasets. At the same scale (1M), our FLAMES Large dataset achieves better MATH5K (68.3), CollegeMath (41.9), GSMPlus (79.4) and average score (61.7) than existing public datasets. On scaling our dataset to 1.5M size (FLAMES XL), we observe drastic improvements as compared to the public datasets across OlympiadBench (+12.8), College Math (+5.8), GSMPlus (+4.9) and Math (+1.3). It is important to note that unlike other synthetic datasets, our FLAMES datasets do not use any proprietary model and rely only on open source models. These findings validate the effectiveness of the FLAMES datasets, showcasing the impact of an optimized mixture of data agents and targeted augmentation in synthetic data generation.

\subsection{Comparison with multiple Student Models}
\label{sec:underlying}

\subfile{../tables/Underlying_Modelsv2.tex}
In this experiment, we compare our FLAMES Large dataset  across several underlying student models. To ensure robust findings, we use four diverse student models --- Qwen2.5-Math-7B and Qwen2.5-14B \cite{yang2024qwen2}, Mathstral-7B \footnote{\url{https://huggingface.co/mistralai/Mathstral-7B-v0.1}}, and Mistral-7B-v0.3 \footnote{\url{https://huggingface.co/mistralai/Mistral-7B-v0.3}}. We compare results with ScaleQuest refreshed as it has best results for public models in Table \ref{tab:open-source}. Table \ref{tab:underlying} shows results for this experiment\footnote{See Table \ref{tab:student} in Appendix \ref{app:open-source} for detailed results.}. We observe that FLAMES Large outperforms refreshed ScaleQuest, achieving drastic gains on competition level benchmarks (+4.8 for DeepSeek-7B, +4.6 for Mathstral-7B, +4.4 for Mistral-7B-v0.3 and +3.7 for Qwen2.5-14B). These results showcase the efficacy of our FLAMES datasets across diverse student models.

\subsection{Comparison of Problem and Solution Generation Models}
\label{sec:synth-model-comparison}
\subfile{../tables/ProbSol_Modelsv3.tex}

For the FLAMES framework we use the Qwen2.5-32B-Instruct and Qwen2.5-Math-7B-Instruct (baseline setting) as the problem generation and solution generation model \cite{yang2024qwen2}, respectively. We now study the impact of updating the problem and solution generation models with weaker DeepSeek models. We generate 50k problems in each setting, using the performant Suggester-Editor agent, and fine-tune the DeepSeek-Math-7B as the student model. Table \ref{tab:probsolmodels} shows results for this study. We observe a big drop (-7.1\% points) in MATH5K numbers on replacing the answer generation model with DeepSeek-Math-7B-RL, but only a small drop (-1.5\% points) on replacing the problem generation model with DeepSeek-v2.5. These results suggest that quality of the solution generation model plays a relatively bigger role than the quality of the problem generation model for math synthetic data pipeline. On the contrary, we observe small improvements (+1.0\% and +2.6\%) in GSM8K numbers. This is likely due to smaller difference in GSM8K numbers for Qwen and DeepSeek, and better alignment of DeepSeek based synthetic data for the DeepSeek-Math-7B student model.



%% file: tables/Quality_Controlv2.tex
\begin{table*}[htb!]
    \centering
    \setlength\tabcolsep{4pt} 
    \resizebox{\textwidth}{!}{ 
    \begin{tabular}{|c|c|c|c|c|c|c|c|c|c|c|c|c|}\hline
    &&&\multicolumn{4}{|c|}{\textbf{Varying Coverage}} &\multicolumn{3}{|c|}{\textbf{Fixed Coverage (45K)}}\\ \hline
    \textbf{Method} &\textbf{Row}&\textbf{Precision} &\textbf{Size} &\textbf{GSM8K} &\textbf{MATH} &\textbf{Avg} &\textbf{GSM8K} &\textbf{MATH} &\textbf{Avg} \\ \hline
    
    Strict Self-Consistency &R1 & Highest &45K (0.3) &82.9 &56.4 &69.7 &82.9 &56.4 &69.7 \\
    Solvability + RM &R2 & High &70K (0.47) &84.5 &57.9 &71.2 &82.9 &56.3 &69.6\\
    Solvability + First & R3& Low &70K (0.47)&84.1 &58.2 &71.2 &83.2 &55.6 &69.4\\
    Majority Self-Consistency & R4& High &90K (0.6) &84.5 &59.4 &72.0 &82.0 &56.1 &69.1\\
    First & R5& Low &\textbf{150K (1.0)}&84.8 &\textbf{\underline{61.1}} &73.0 &\textbf{\underline{83.4}} &55.8 &69.6\\ 
    Majority + First & R6& Medium &\textbf{150K (1.0)}&\textbf{\underline{86.4}} &60.9 &\textbf{\underline{73.7}} &83.0 &\textbf{\underline{56.6}} &\textbf{\underline{69.8}}\\ \hline
    \end{tabular}
    }
\caption{Results of synthetic data quality control strategies for two settings of \textit{Varying Coverage} and \textit{Fixed Coverage}. Size column refers to number of problems remaining after filtering followed by fraction of actual in parenthesis. Each result is using DeepSeek-Math-7B fine-tuned on problems synthesized using Suggester-Editor agent.}
\label{tab:quality-control}
\end{table*}

%% file: tables/Agents.tex
\begin{table*}[ht]
    \centering
    \setlength\tabcolsep{4pt} 
    \resizebox{\textwidth}{!}{ 
    \begin{tabular}{|c|c|c|c|c|c|c|c|c|}\hline
        & \multicolumn{3}{|c|}{\textbf{In-Domain}} & \multicolumn{1}{|c|}{\textbf{OOD}} & \multicolumn{2}{c|}{\textbf{Robustness}} & \multicolumn{1}{c|}{\textbf{Competition}} & \\ \hline
        \textbf{Model} & \textbf{GSM8K} & \textbf{MATH5K} & \textbf{Average} & \textbf{College Math} & \textbf{Distraction} & \textbf{GSMPlus} & \textbf{Olympiad Bench} & \textbf{Average} \\ \hline
        DeepSeek-Math-7B (Base) & 64.2 & 36.2 & 50.2 & 15.9 & 17.6 & 22.6 & 5.3 & 28.8 \\
        DeepSeek-Math-7B-Instruct & 82.9 & 51.7 & 67.3 & 33.9 & 60.0 & 70.1 & 13.8 & 50.5 \\ \hline
        \textbf{Agents} & \multicolumn{8}{|c|}{\textbf{In-Domain Practice}} \\ \hline
        Seeded Key Concepts & 81.7 & 58.7 & 70.2 & 38.6 & 67.8 & 70.9 & 22.4 & 54.5 \\
        Key Concepts & 79.2 & 56.2 & 67.7 & 38.6 & 66.1 & 69.4 & 22.7 & 53.2 \\
        Paraphrase & 81.2 & 57.5 & 69.3 & 38.0 & 62.6 & 69.7 & 22.5 & 53.8 \\
        Few-Shot & 81.9 & 57.2 & 69.6 & 37.2 & 66.2 & 70.1 & 23.1 & 53.9 \\ \hline
        \textbf{Agents} & \multicolumn{8}{|c|}{\textbf{In-Domain Complexity Enhancing} } \\ \hline
        IQC & \underline{86.0} & 59.9 & 73.0 & 38.7 & 72.1 & 75.7 & 24.4 & 56.9 \\
        Suggester-Editor & 85.3 & \textbf{\underline{61.0}} & \textbf{\underline{73.2}} & 39.4 & 72.1 & \textbf{\underline{75.9}} & 25.2 & 57.4 \\ \hline
        \textbf{Agents} & \multicolumn{8}{|c|}{\textbf{Out-of-Domain Enhancing}} \\ \hline
        QFT & 79.0 & 57.5 & 68.3 & 40.5 & 67.0 & 68.6 & 23.6 & 53.8 \\ \hline
        \textbf{Agents} & \multicolumn{8}{|c|}{\textbf{Robustness Enhancing}} \\ \hline
        Ask Me Anything & 83.2 & 56.4 & 69.8 & 40.0 & 65.1 & 71.4 & 23.7 & 54.9 \\
        FOBAR & 80.5 & 55.8 & 68.2 & 37.9 & 61.2 & 69.6 & 19.9 & 52.7 \\
        Self-Verification & 82.7 & 57.9 & 70.3 & 38.1 & 60.9 & 71.7 & 23.4 & 54.8 \\ \hline
        \textbf{Agents} & \multicolumn{8}{|c|}{\textbf{Novel Agents}} \\ \hline
        Distraction Insertion (Ours) & 83.3 & 59.4 & 71.3 & 39.6 & \textbf{\underline{72.4}} & 73.5 & 24.7 & 56.1 \\
        Taxonomy Key Concepts (Ours) & 77.1 & 56.1 & 66.6 & 40.9 & 64.7 & 67.3 & 21.8 & 52.6 \\ \hline
        \textbf{Dataset} & \multicolumn{8}{|c|}{\textbf{FLAMES Data Mixture}} \\ \hline
        FLAMES Small &85.2 &60.0 &72.6 &\textbf{\underline{41.4}} &72.2 &74.7 &\textbf{\underline{26.1}} & \textbf{\underline{57.5}} \\ \hline
    \end{tabular}}
    \caption{Results comparing $10$ existing and $2$ novel data agents. We also include results for FLAMES\_small (150K) for easy comparison with all agents. DeepSeek-Math-7B is fine-tuned using 150K problems for each setting. }
    \label{tab:agents}
\end{table*}


%% file: tables/OpenSourceComp_Short.tex
\begin{table*}[ht]
    \centering
    \setlength\tabcolsep{4pt} 
    \resizebox{\textwidth}{!}{ 
    \begin{tabular}{|c|c|c|c|c|c|c|c|c|}\hline
        & &\multicolumn{2}{|c|}{\textbf{In-Domain}} &\textbf{OOD} &\multicolumn{2}{|c|}{\textbf{Robustness}} &\textbf{Competition} & \\\hline
        \multicolumn{2}{|c|}{\textbf{Model}} &\textbf{GSM8K} &\textbf{MATH5K} &\textbf{College Math} &\textbf{Distraction} &\textbf{GSMPlus} &\textbf{Olympiad Bench} &\textbf{Average} \\ \hline
        \multicolumn{2}{|c|}{DeepSeek-Math-7B (Base)} &64.2 &36.2 &15.9 &17.6 &22.6 &5.3 &28.8 \\
        \multicolumn{2}{|c|}{DeepSeek-Math-7B-Instruct} &82.9 &51.7 &33.9 &60 &70.1 &13.8 &50.5 \\ \hline
        \textbf{Dataset} &\textbf{Dataset Size} &\multicolumn{7}{|c|}{\textbf{Existing Math Datasets}} \\ \hline
        GSM8K/MATH &15K &70.1 &34.9 &33.9 &60 &70.1 &13.8 &44.6 \\
        NuminaMath &860K &81.4 &53.1 &36.6 &61.9 &70.2 &20.4 &52.3 \\
        OpenMathInstruct2 (R) &600K &84.5 &66.4 &40.7 &64.7 &72.9 &30.5 &59.0 \\
        MetaMathQA (R) &150K &85.0 &55.8 &36.7 &64.1 &72.9 &21.6 &54.4 \\
        OrcaMath (R) &200K &84.5 &52.8 &36.9 &68.1 &74.9 &19.7 &53.8 \\
        MMIQC (R) &220K &84.8 &61.5 &38.4 &63.7 &73.6 &24.4 &56.5 \\
        ScaleQuest (R) &1M &\textbf{\underline{89.4}} &66.0 &41.4 &71 &78.3 &25 &60.0 \\ \hline
        \textbf{Dataset} &\textbf{Dataset Size} &\multicolumn{7}{|c|}{\textbf{FLAMES Data}} \\ \hline
        FLAMES Large &1M &89.2 &\textbf{\underline{68.3}} &41.9 &76.1 &79.4 &29.8 &61.7 \\
        FLAMES XL &1.5M &87.7 &67.7 &\textbf{\underline{47.2}} &\textbf{\underline{80.3}} &\textbf{\underline{83.2}} &\textbf{\underline{43.3}} & \textbf{\underline{65.8}} \\
        \hline
    \end{tabular}}
    \caption{Comparison of FLAMES datasets with open-source math reasoning datasets. DeepSeek-Math-7B is fine-tuned using unique problems for each dataset, alongside refreshed (R) solutions using Qwen2.5-Math-7B-Instruct for fair comparison.}
    \label{tab:open-source}
\end{table*}

%% file: tables/Underlying_Modelsv2.tex
\begin{table}[ht]
   \centering
    \setlength\tabcolsep{4pt} 
    \resizebox{0.48\textwidth}{!}{    
    \begin{tabular}{|c|c|c|c|c|c|}\hline
    & & \multicolumn{2}{c|}{\textbf{In-Domain}}& \multicolumn{1}{c|}{\textbf{Competition}}& \textbf{Avg}\\ \hline
    \textbf{Student Model}&\textbf{Dataset}&\textbf{GSM8K}&\textbf{MATH}&\textbf{\makecell{Olympiad \\ Bench}}&\\ \hline
    DeepSeek-7B&ScaleQuest (R)&\textbf{\underline{89.4}}&66.0&25.0&60.0\\
    DeepSeek-7B&FLAMES (L)&89.2&\textbf{\underline{68.3}}&\textbf{\underline{29.8}}& \textbf{\underline{61.7}} \\ \hline
    Qwen2.5M-7B&ScaleQuest (R)&\textbf{\underline{93.9}}&80.7&\textbf{\underline{41.3}}&69.2 \\
    Qwen2.5M-7B&FLAMES (L)&93.5&\textbf{\underline{81.4}}&40.9&\textbf{\underline{69.5}} \\ \hline
    Mathstral-7B&ScaleQuest (R)&88.4&65.7&24.6&59.0 \\
    Mathstral-7B&FLAMES (L)&\textbf{\underline{89.2}}&\textbf{\underline{68.1}}&\textbf{\underline{29.2}}&\textbf{\underline{61.1}} \\ \hline
    Mistral-7B&ScaleQuest (R)&84.7&59.3&19.7&54.8 \\
    Mistral-7B&FLAMES (L)&\textbf{\underline{86.4}}&\textbf{\underline{63.6} }& \textbf{\underline{24.1}}&\textbf{\underline{57.5}} \\ \hline
    Qwen2.5-14B&ScaleQuest (R)&93.1&75.6&34.2&66.2\\
    Qwen2.5-14B&FLAMES (L)& \textbf{\underline{93.3}}& \textbf{\underline{76.7}}& \textbf{\underline{37.9}}&\textbf{\underline{67.4}}\\
    \hline
    \end{tabular}
    }
    \caption{Comparison of FLAMES\_large and ScaleQuest across diverse student models. DeepSeek-7B refers to DeepSeek-Math-7B, Mistral-7B refers to Mistral-7B-v0.3, Qwen2.5M-7B refers to Qwen2.5-Math-7B, FLAMES (L) refers to FLAMES\_large, ScaleQuest (R) refers to ScaleQuest problems with Qwen2.5-Math-7B-Instruct solutions. Both datasets contain 1M problems.}
    \label{tab:underlying}
\end{table}

%% file: tables/ProbSol_Modelsv3.tex

\begin{table}
    \centering
    \resizebox{0.50\textwidth}{!}{
        \begin{tabular}{|c|c|c|c|c|c|c|} \hline
        \textbf{\makecell{Problem \\Generation}}&\textbf{\makecell{Solution \\Generation}}&\textbf{GSM8K}&\textbf{MATH}&\textbf{Average} \\ \hline
        Qwen2.5-32B&Qwen2.5-7B&82.5&\textbf{\underline{56.4}}&\textbf{\underline{69.5}} \\
        DeepSeek-v2.5&Qwen2.5-7B&83.5&54.9&69.2 \\
        Qwen2.5-32B&DeepSeek-7B&\textbf{\underline{85.1}}&49.3&67.2 \\\hline
        \end{tabular}
    }
\caption{Results of varying problem and solution generation models. Qwen2.5-32B refers to Qwen2.5-32B-Instruct model, DeepSeek-7B refers to DeepSeek-Math-7B-RL. Finetuning is done with 50K Suggester-Editor problems for DeepSeek-Math-7B as student model.}
\label{tab:probsolmodels}
\end{table}

%% file: sections/08_Conclusion.tex
\section{Conclusion}
We introduced the FLAMES framework, which enables fine-grained analysis of the math data synthesis pipeline. We performed a controlled study of $12$ data synthesis strategies (including $2$ novel strategies), $6$ data quality control strategies, $2$ problem generation models, and $2$ solution generation models, providing valuable insights for improving LLM math reasoning with synthetic data. We studied the impact of mixing data from different agents and designed an effective blend (the FLAMES datasets) without use of any proprietary models. Our rigorous evaluations establish drastic improvements with our proposed FLAMES datasets over public math datasets.


%% file: sections/09_Limitations.tex
\section{Limitations and Potential Risks}
One limitation of our work is reliance on a strong teacher model for solution generation. This assumption may be violated when working with less common languages which lack sufficient math data to begin with. While this may be mitigated by translating our English math data to those languages, we leave study of other languages for future work.

We acknowledge that there is a potential risk of inherent bias in our generated data due to bias in the problem and solution generating models or the original seed data \cite{lebrun2022evaluating, yu2023large}. This may be alleviated by introducing an additional step in the FLAMES framework for detecting and filtering biased samples, however we leave this for future work.

%% file: sections/AA_Appendixv2.tex

\section*{Contents of Appendix}

\startcontents[sections]
\printcontents[sections]{l}{1}{\setcounter{tocdepth}{2}}

\section{FLAMES Framework Details}
\label{app:framework-details}

\subfile{../tables/framework_details.tex}

Figure \ref{fig:overview} shows a flow diagram of the FLAMES framework. Broadly, the synthetic data pipeline can be considered as consisting of the following steps. 1) Generating synthetic problems with a problem synthesis LLM and some seed problems (or taxonomy) according to some data agent, 2) generating solutions for synthetic problems with a solution synthesis LLM, 3) filtering the problems and solutions for quality control, 4) training a student model with this synthetic dataset, and 5) evaluating the student model on multiple evaluation benchmarks. There are multiple factors (or design decisions) that play key role in final student model performance with synthetic data. As part of the FLAMES framework, we list these factors. We include a table of these factors, their values in the FLAMES framework, and whether we choose values based on a controlled experiment or prior work in Table \ref{tab:app:framework-details}.

\textbf{Problem Generation:} Seeded problem generation agents in the FLAMES framework utilize the train splits from the popular GSM8K \cite{cobbe2021training} and MATH \cite{hendrycks2measuring} training sets. These agents interact with the Qwen2.5-32B-Instruct model \cite{yang2024qwen2} to generate new problems, an open-source instruction-tuned model which achieves 95.9 on GSM8K and 83.1 on MATH5K. We show in Section \ref{sec:synth-model-comparison} that this model leads to better performance than the larger open-source DeepSeek-v2.5 model \cite{deepseekv2}, while being more computationally efficient. Problems are generated using a temperature of 0.7 with sampling hyperparameters of $top\_p = 0.9$, $top\_k = 50$, and a repetition penalty of 1 for a maximum of 2,048 new tokens.

\textbf{Solution Generation:} Since solution generation can be less efficient due to higher inference time compute, we choose to use the smaller Qwen2.5-Math-7B-Instruct model \cite{yang2024qwen2} for solution synthesis in the FLAMES framework. This follows \cite{scalequest}, who use the earlier Qwen2-Math-7B-Instruct \cite{yang2024qwen2_old} for both problem and solution synthesis. Solution generation uses the same generation parameters as in problem synthesis. As we show in Section \ref{sec:quality-control-exp}, taking the first solution generated by Qwen2-Math-7B-Instruct for each synthetic problem leads to good performance while significantly reducing required generation compute.

\textbf{Quality Control:} All problems generated in the FLAMES framework are first deduplicated using exact match. While we do not use test splits of GSM8K or MATH when synthesizing problems, we exercise caution and follow \cite{orcamath} in decontaminating synthetic problems against the GSM8K and MATH test sets. We remove any synthetic problem if there is a high amount of token overlap between the synthetic problem and a test set problem. In other words, if 95\% of the 8-grams in a test set problem are generated in a synthetic problem, we remove that synthetic problem. 

\textbf{Dataset Size:} Using these problem and solution generation processes, we generate 150k problems \textit{after problem deduplication and decontamination} for each agent in the FLAMES framework. In other words, we fix the size of the SFT dataset for each data agent at 150k. Of these 150k problems, 75k are generated using GSM8K seed problems and 75k are generated using MATH seed problems\footnote{FOBAR and Self-Consistency agents use fixed values in existing problems, and are limited in the number of unique problems which they generate. For these, we include as many unique problems as are available.}. Due to their reliance on seed problem permutations and the limited size of the human-crafted training datasets in GSM8K and MATH, the FOBAR and Self-Verification agents are limited in number of unique synthetic problems they generate.

\textbf{Training Details:} We use the DeepSeek-Math-7B model \cite{shao2024deepseekmath} as the student model in the FLAMES framework. We do full-parameter fine-tuning on Amazon EC2 P4 instances\footnote{\url{https://aws.amazon.com/ec2/instance-types/p4/}} using Deepspeed Zero3 \cite{rajbhandari2020zeromemoryoptimizationstraining} for distributed training across 8 A100 GPUs. We train using a batch size of 4 for 5 epochs, saving 10 total checkpoints. We implement training using the SWIFT library\footnote{\url{https://github.com/modelscope/ms-swift}} \cite{zhao2024swiftascalablelightweightinfrastructure}, using default hyperparameters. 

\textbf{Evaluation:} We evaluate fine-tuned student models using the Qwen2.5-Math evaluation framework\footnote{\url{https://github.com/QwenLM/Qwen2.5-Math}} \cite{yang2024qwen2}. We report scores of the checkpoint with the highest average score on GSM8K and MATH. All models are evaluated using the same system prompt recommended by the Qwen2.5-Math framework (see Appendix \ref{app:agent-prompts} for the system prompt). The evaluation decoding was done with temperature 0, allowing for a maximum of 2,048 new tokens. Answer extraction and matching is handled by the Qwen2.5-Math framework, which is borrowed from the release of the MATH dataset \cite{hendrycks2measuring}.

\section{Related Works}

\subsection{LLM Math Reasoning}
The ability to solve math word problems is considered a key measure of large language model performance \cite{cobbe2021training, hendrycks2measuring}. Recent large language models have used math reasoning as an important metric to highlight model reasoning capability \cite{hendrycksmeasuring, team2024gemma, dubey2024llama, yang2024qwen2, deepseekai2025deepseekr1incentivizingreasoningcapability}. Various strategies have been proposed to improve large language model performance on math reasoning, including instruction tuning \cite{luo2023wizardmath, scalequest}, prompting techniques \cite{wei2022chain, chia2023contrastivechainofthoughtprompting}, and solution construction \cite{omi1}. Our work focuses on evaluating problem synthesis techniques for improving LLM math reasoning. 

\subsection{Data Synthesis for LLM Math Reasoning}
Existing works in evaluating synthetic data agents for improving LLM math reasoning have shown that various methods for generating synthetic math word problems can improve downstream performance of underlying student models fine-tuned on synthetic problems \cite{chen2025advancingmathreasoninglanguage, kim2024evaluatinglanguagemodelssynthetic, hase2024unreasonableeffectivenesseasytraining, he2024guidingcomplexitymakesgood, lu2024mathgenie}. MetaMathQA \cite{metamath} rewrites questions from multiple perspectives, enabling augmentation without additional knowledge. Open-Math-Instruct-2 uses few-shot LLM prompting to generate novel problems \cite{omi2}. Other works introduce multi-step approaches to synthesizing questions \cite{orcamath, luo2023wizardmath, mmiqc}.

\section{Data Synthesis Agents for LLM Math Reasoning}
\label{app:agent-details}

\subsection{Agent Descriptions}

\subfile{../tables/agent_list.tex}

We now categorize 4 types of approaches and detail the 10 existing agents compared in the FLAMES framework (see Section \ref{sec:data-agents}). For reference, we include citations and agent types in Table \ref{tab:app:agent-list}.

\textbf{In-Domain Practice:} In-domain practice agents are designed to increase the amount of in-domain data available for model training. The goal of in-domain practice agents is to synthesize a large amount of math word problems which are similar in complexity, topic, and grade-level to problems in the training data. The \textbf{Few-Shot} agent, proposed in \cite{omi2}, uses existing problems as in-context examples to generate a novel problem. The \textbf{Paraphrasing} agent (\cite{metamath}) paraphrases existing problems. \cite{kpdds} introduce the use of key concepts as an intermediate problem representation for synthesizing novel problems. We evaluate a \textbf{Key Concepts} agent, which first extracts key concepts from an existing problem, and then uses those key concepts to generate novel problems. The related \textbf{Seeded Key Concepts} agent uses both extracted key concepts and the original existing problem to guide the generation of novel problems.

\textbf{In-Domain Complexity Enhancing:} There have been several proposed agents which aim to increase the complexity of existing math word problems. We select two recent agents to evaluate this type of strategy in our work, although other similar strategies have been proposed including Evol-Instruct \cite{luo2023wizardmath, li2024mugglemath}.  The \textbf{Suggester-Editor} agent first suggests edits for problems to introduce additional reasoning steps, then synthesizes a novel problem by making those edits \cite{orcamath}. The \textbf{Iterative Question Composing (IQC)} agent turns an existing problem into a subproblem of a more complex synthetic problem \cite{mmiqc}. Both the Suggester-Editor and IQC agents may repeat their processes multiple times.

\textbf{Robustness Enhancing:} Some proposed agents aim to alter existing problems to increase the robustness of the student model's reasoning process. These agents include the \textbf{Ask Me Anything} agent \cite{orcamath}, which converts an existing problem and solution into a statement, then prompts the teacher model to ask questions about the statement. Similarly, the \textbf{Self-Verification} agent \cite{metamath} converts an existing problem and solution into a statement, then replaces a number in the problem with a variable and asks the model to reason backwards to obtain the value of the variable. The \textbf{FOBAR} agent \cite{metamath} is similar to Self-Verification, except instead of having the teacher model rewrite the problem and solution into a statement before variable replacement, there is a template sentence added to the end of the problem which includes the problem's answer and the resulting question\footnote{See Appendix \ref{app:agent-prompts} for specific prompts}.

\textbf{Out-of-Domain Enhancing:} Recently, \cite{scalequest} proposed a method for synthesizing math reasoning data by using only a ``small-size'' (e.g. 7B) open-source model without a complex augmentation strategy. Their method revolves around the \textbf{Question Fine-Tuning (QFT)} process, which lightly fine-tunes a 7B math specialist model on the GSM8K and MATH training datasets, then prompts the fine-tuned model with only the system prompt to synthesize novel math word problems. We replicate their proposed process using Qwen2.5-Math-7B-Instruct \cite{yang2024qwen2} to create the QFT agent, designed to enhance general math reasoning performance. We call this an out-of-domain agent, as the agent prompt is not seeded with any specific in-domain problem.

\begin{figure}[t]
 \centering
  \includegraphics[width=\columnwidth]{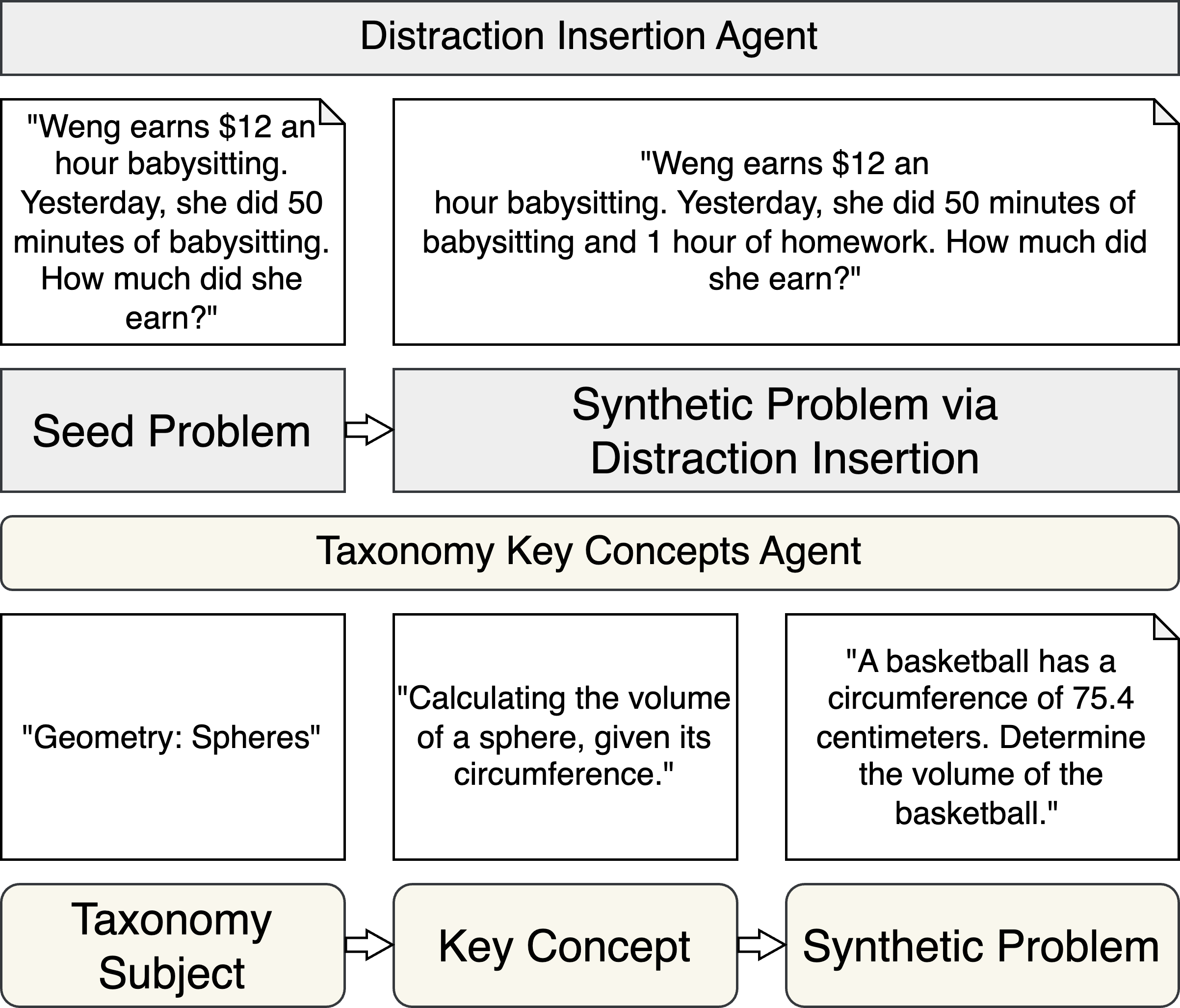}
  \caption{Examples of our novel Distraction Insertion and Taxonomy-Based Key Concepts data agents.}
  \label{fig:agents}
\end{figure}

\textbf{Novel Agents} are described in Section \ref{sec:data-agents}. Examples of synthetic problems generated by novel agents can be seen in Figure \ref{fig:agents}.

\subsection{Agent Prompt Locations}
\label{sec:promptlocation}
\subfile{../tables/agent_prompt_locations.tex}

In Table \ref{tab:app:agent-prompts} we provide the locations of each prompt used in the FLAMES framework. Prompts for the novel Distraction Insertion agent can be found in Appendix \ref{app:agent-prompts}.

\section{Additional Experiments and Detailed Results}

\subsection{Comparison with Open-Source Synthetic Datasets}
\label{app:open-source}

\subfile{../tables/OpenSourceComp.tex}
\subfile{../tables/Underlying_Models.tex}

Full results from Section \ref{sec:os-dataset-exp} are given in Table \ref{tab:open-source-full} and Section \ref{sec:underlying} are given in Table \ref{tab:student}. 


\subsection{Combining Data Generation Agents}
\label{app:combining-agents}

\subfile{../tables/DataMixture.tex}
In this section, we discuss our study of mixing data from different agents (Table \ref{tab:mixture}). \underline{First}, we analyze if mixing Taxonomy Key Concepts agent (having best out-of-domain performance) can improve out-of-domain reasoning for the best complexity-enhancing Suggester \& Editor agent. By mixing data from Suggester \& Editor and Taxonomy Key Concepts agents in equal proportion (mixture-A), we observe improved CollegeMath score of 41.6 exceeding both the agents. However, we observe slight drop in other metrics as compared to the Suggester \& Editor agent. \underline{Second}, we investigate impact of different mixing proportions (mixture-A, mixture-B and mixture-C) for Suggester \& Editor and Taxonomy Key Concept agent. We observe that mixture-C (75/25 split) yields a better trade-off\footnote{We observe drop in Competition level metrics for mixture-C, but we recover that with our mixture-D} as there is big drop of In-Domain metrics for mixture-A and OOD metrics for mixture-B. 

\underline{Third}, we investigate whether enhancing diversity by introducing a different complexity-enhancing IDC agent further improves the performance. For mixture-D, we keep same 75/25 proportion of complexity-enhancing and out-of-domain enhancing, by reducing Suggester \& Editor proportion to 50\% and including 25\% of IQC. With mixture-D, we recover OlympiadBench performance with slight drop in OOD metrics as compared to mixture-C. \underline{Fourth}, we explore whether adding small proportion (10\%) of robustness-focused agents improves overall performance. We observe that including Distraction Insertion (mixture FLAMES\_Small) further boosts performance, leading to a mixture with best GSM8K (85.2), Math (60.0), OlympiadBench (26.1) and Average score (57.5). These findings establish that mixing our novel Distraction Insertion agent data can bring significant improvement to the student model’s performance.

With our FLAMES\_Small mixture, we observe balanced performance across the five evaluation datasets, with performance surpassing all $12$ agents on CollegeMath and Olympiad-Bench.

\subsection{Solvability Filtering}
\label{app:solvability}

\subfile{../tables/Solvability.tex}

We find in Section \ref{sec:quality-control-exp} that the recently-proposed solvability filtering strategy \cite{scalequest} for quality control of synthetic math problems does not lead to performance gains from fine-tuned student models. We hypothesize that this is because the solvability filtering strategy also removes solvable problems from the synthetic data pool. To test this hypothesis we use the human-crafted MATH500 \cite{hendrycks2measuring} dataset, filtering for solvability. We observe in Table \ref{tab:app:solvability} that the solvability filter removes 30\% of the real problems, with a higher percentage of more difficult problems being removed (e.g. 50\% of level 5 problems). These results indicate that solvability filtering is unreliable.

\section{Additional Information}

\begin{figure*}[t]
 \centering
  \includegraphics[width=1.8\columnwidth]{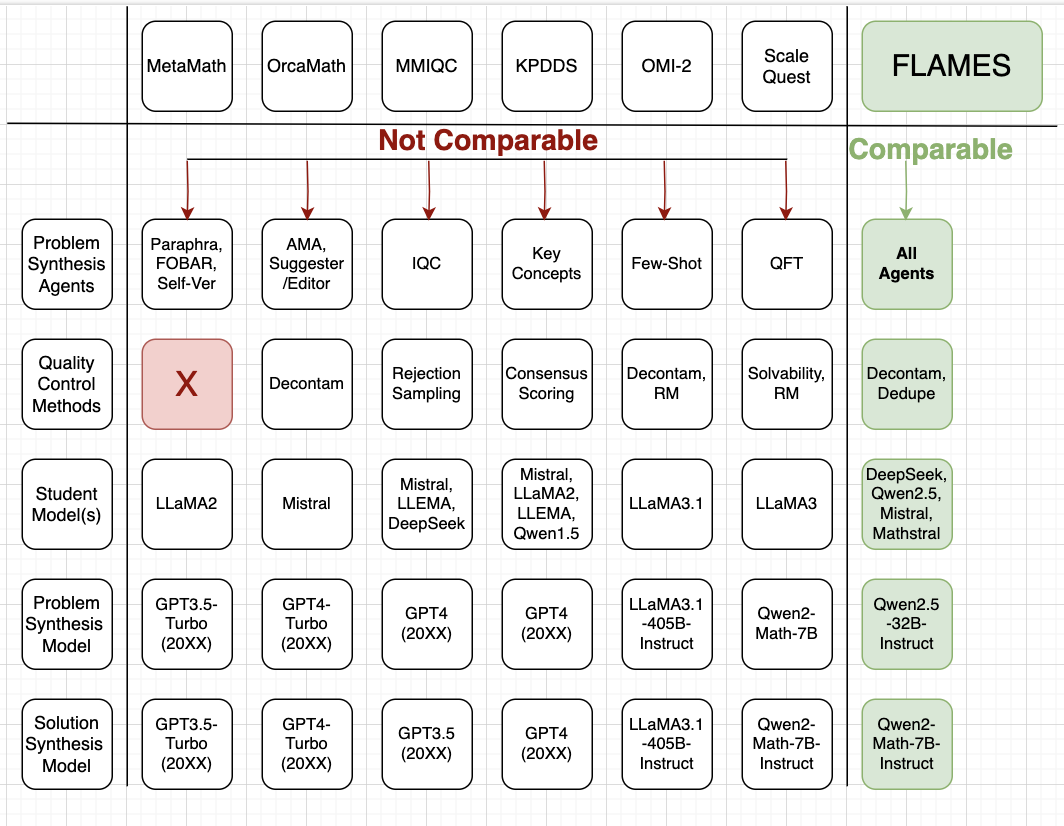}
  \caption{The full landscape of several recently-proposed math data synthesis works. Each work uses different synthesis models, student models, and quality control measures, making comparison of problem synthesis agents impractical.}
  \label{fig:comparison_appendix}
\end{figure*}

\underline{Landscape of recent Math data synthesis:} We include a larger version of Figure \ref{fig:preview} for reference in Figure \ref{fig:comparison_appendix}, further highlighting the need for the unified math data synthesis evaluation environment provided by the FLAMES framework.

\subfile{../tables/Appendix_Models.tex}
\subfile{../tables/Appendix_Datasets.tex}
\subfile{../tables/Appendix_Experiments.tex}

\underline{Table of models and datasets used}: We also include tables giving an overview of all models and datasets used, then a table of all experiments and the locations of their corresponding results. Table \ref{tab:app:models} details all models used in the FLAMES framework, as well as models used in our controlled experiments. Table \ref{tab:app:datasets} details all datasets used in the FLAMES framework. Table \ref{tab:app:experiments} details all controlled experiments run in this paper, along with their sections and results tables.

\subfile{AB_Prompt_Appendix.tex}

%% file: tables/framework_details.tex
\begin{table*}[ht]
    \centering
    \setlength\tabcolsep{4pt} 
    \resizebox{\textwidth}{!}{ 
    \begin{tabular}{|c|c|c|c|}\hline
        \textbf{Factor} &\textbf{Value} &\textbf{Controlled/Prior} &\textbf{Location} \\\hline
        \multicolumn{4}{|c|}{\textbf{Problem Generation}} \\ \hline
        Data Synthesis Agent &10 Existing, 2 Novel &Controlled &Section \ref{sec:data-agents-exp} \\ \hline
        Problem Generation Model &Qwen2.5-32B-Instruct &Controlled &Section \ref{sec:synth-model-comparison} \\ \hline
        Problem Generation Decoding Parameters &See Appendix \ref{app:framework-details} &Prior Work &Appendix \ref{app:framework-details} \\ \hline
        Deduplication &Exact Match &Prior Work &Section \ref{sec:framework} \\ \hline
        Benchmark Decontamination &N-Gram Overlap &Prior Work &Section \ref{sec:framework} \\ \hline
        \multicolumn{4}{|c|}{\textbf{Solution Generation}} \\ \hline
        Solution Generation Model &Qwen2.5-Math-7B-Instruct &Controlled &Section \ref{sec:synth-model-comparison} \\ \hline
        Solution Generation Decoding Parameters &See Appendix \ref{app:framework-details} &Prior Work &Appendix \ref{app:framework-details} \\ \hline
        Solution Sampling Strategy &First Solution &Controlled &Section \ref{sec:quality-control-exp} \\ \hline
        Solution Verification Strategy &None &Controlled &Section \ref{sec:quality-control-exp} \\ \hline
        \multicolumn{4}{|c|}{\textbf{Supervised Fine-Tuning}} \\ \hline
        Student Model &\makecell{DeepSeek-Math-7B, Qwen2.5-Math-7B, \\ Mathstral-7B, Mistral-7B-v0.3, \\ Qwen2.5-14B} &Controlled &Section \ref{sec:underlying} \\ \hline
        Training Hyperparameters &See Section \ref{sec:framework} &Prior Work &Section \ref{sec:framework} \\ \hline
        Training Setup &See Section \ref{sec:framework} &Prior Work &Section \ref{sec:framework} \\ \hline
        \multicolumn{4}{|c|}{\textbf{Evaluation}} \\ \hline
        Checkpoint Selection &See Appendix \ref{app:framework-details} &Prior Work &Appendix \ref{app:framework-details} \\ \hline
        Evaluation Prompts &See Appendix \ref{app:framework-details} &Prior Work &Appendix \ref{app:framework-details} \\ \hline
        Evaluation Decoding Parameters &See Appendix \ref{app:framework-details} &Prior Work &Appendix \ref{app:framework-details} \\ \hline
        Answer Extraction \& Matching &See Appendix \ref{app:framework-details} &Prior Work &Appendix \ref{app:framework-details} \\ \hline
        \end{tabular}
}
    \caption{Values of all factors fixed in the FLAMES framework. Each value is chosen based on either controlled experiments of prior work. Locations of framework factor details, and their related controlled experiments, are also given.}
    \label{tab:app:framework-details}
\end{table*}

%% file: tables/agent_list.tex
\begin{table}
    \centering
    \resizebox{0.45\textwidth}{!}{
        \begin{tabular}{|c|c|}\hline
        \textbf{Agent} &\textbf{Work} \\\hline
        \multicolumn{2}{|c|}{\textbf{In-Domain Practice}} \\ \hline
        Few-Shot &\cite{omi2} \\ \hline
        Paraphrasing &\cite{metamath} \\ \hline
        Key Concepts &\cite{kpdds} \\ \hline
        Seeded Key Concepts &\cite{kpdds} \\ \hline
        \multicolumn{2}{|c|}{\textbf{In-Domain Complexity Enhancing}} \\ \hline
        Suggester-Editor &\cite{orcamath} \\ \hline
        IQC &\cite{mmiqc} \\ \hline
        \multicolumn{2}{|c|}{\textbf{Robustness Enhancing}} \\ \hline
        Ask Me Anything &\cite{orcamath} \\ \hline
        Self-Verification &\cite{metamath} \\ \hline
        FOBAR &\cite{metamath} \\ \hline
        Distraction Insertion &Our \\ \hline
        \multicolumn{2}{|c|}{\textbf{Out-Of-Domain Enhancing}} \\ \hline
        Question Fine-Tuning (QFT) &\cite{scalequest} \\ \hline
        Taxonomy Key Concepts &Our \\
        \hline 
\end{tabular}
    }
\caption{List of all agents evaluated in the FLAMES framework.}
\label{tab:app:agent-list}
\end{table}

%% file: tables/agent_prompt_locations.tex
\begin{table}
    \centering
    \resizebox{0.45\textwidth}{!}{
        \begin{tabular}{|c|c|} \hline
            \textbf{Agent} &\textbf{Prompt Location} \\ \hline
            \multicolumn{2}{|c|}{\textbf{In-Domain Practice}} \\ \hline
            Few-Shot &\cite{omi2} Appendix D.2 \\ \hline
            Paraphrasing &\cite{metamath} Section 3.2 \\ \hline
            Key Concepts &Appendix \ref{app:agent-prompts} \\ \hline
            Seeded Key Concepts &Appendix \ref{app:agent-prompts} \\ \hline
            \multicolumn{2}{|c|}{\textbf{In-Domain Complexity Enhancing}} \\ \hline
            Suggester-Editor &\cite{orcamath} Section 2 \\ \hline
            IQC &\cite{mmiqc} Figures 5, 6 \\ \hline
            \multicolumn{2}{|c|}{\textbf{Robustness Enhancing}} \\ \hline
            Ask Me Anything &\cite{orcamath} Section 2 \\ \hline
            Self-Verification &\cite{metamath} Section 3.3 \\ \hline
            FOBAR &\cite{metamath} Section 3.3 \\ \hline
            \multicolumn{2}{|c|}{\textbf{Novel Agents}} \\ \hline
            Taxonomy Key Concepts & \makecell{Prompts: \cite{nvidia2024nemotron4340btechnicalreport} \\ Appendix B.5, Taxonomy: Section \ref{sec:data-agents}} \\ \hline
            Distraction Insertion &Appendix \ref{app:agent-prompts} \\
            \hline
        \end{tabular}
    }
\caption{Locations of prompts used for agents in the FLAMES framework.}
\label{tab:app:agent-prompts}
\end{table}

%% file: tables/OpenSourceComp.tex
\begin{table*}[ht]
    \centering
    \setlength\tabcolsep{4pt} 
    \resizebox{\textwidth}{!}{ 
    \begin{tabular}{|c|c|c|c|c|c|c|c|c|}\hline
        & &\multicolumn{2}{|c|}{\textbf{In-Domain}} &\textbf{OOD} &\multicolumn{2}{|c|}{\textbf{Robustness}} &\textbf{Competition} & \\\hline
        \multicolumn{2}{|c|}{\textbf{Model}} &\textbf{GSM8K} &\textbf{MATH5K} &\textbf{College Math} &\textbf{Distraction} &\textbf{GSMPlus} &\textbf{Olympiad Bench} &\textbf{Average} \\ \hline
        \multicolumn{2}{|c|}{DeepSeek-Math-7B (Base)} &64.2 &36.2 &15.9 &17.6 &22.6 &5.3 &28.8 \\
        \multicolumn{2}{|c|}{DeepSeek-Math-7B-Instruct} &82.9 &51.7 &33.9 &60 &70.1 &13.8 &50.5 \\ \hline
        \textbf{Dataset} &\textbf{Dataset Size} &\multicolumn{7}{|c|}{\textbf{Original Dataset}} \\ \hline
        GSM8K/MATH &15K &70.1 &34.9 &33.9 &60 &70.1 &13.8 &44.6 \\
        NuminaMath &860K &81.4 &53.1 &36.6 &61.9 &70.2 &20.4 &52.3 \\
        MetaMathQA &400K &78.5 &41.6 &30.9 &53.4 &65.3 &9.9 &45.2 \\
        OrcaMath &200K &77.9 &42.3 &29.1 &63.4 &67.8 &13.5 &46.1 \\
        OpenMathInstruct2 &600K &83.5 &55.3 &36.8 &64.4 &72.3 &19.6 &53.5 \\
        MMIQC &1M &77.7 &42.3 &31.6 &53.8 &63.4 &11.9 &45.4 \\
        ScaleQuest &1M &86.4 &64.6 &42.7 &71.1 &76.7 &27.6 &59.6 \\ \hline
        \textbf{Dataset} &\textbf{Dataset Size} &\multicolumn{7}{|c|}{\textbf{Refreshed with Qwen2.5-Math-7B solutions}} \\ \hline
        OpenMathInstruct2 &600K &84.5 &66.4 &40.7 &64.7 &72.9 &30.5 &59.0 \\
        MetaMathQA &150K &85.0 &55.8 &36.7 &64.1 &72.9 &21.6 &54.4 \\
        OrcaMath &200K &84.5 &52.8 &36.9 &68.1 &74.9 &19.7 &53.8 \\
        MMIQC &220K &84.8 &61.5 &38.4 &63.7 &73.6 &24.4 &56.5 \\
        ScaleQuest &1M &\underline{89.4} &66.0 &41.4 &71 &78.3 &25 &60.0 \\ \hline
        \textbf{Dataset} &\textbf{Dataset Size} &\multicolumn{7}{|c|}{\textbf{FLAMES Data}} \\ \hline
        FLAMES\_large &1M &89.2 &\underline{68.3} &41.9 &76.1 &79.4 &29.8 &61.7 \\
        FLAMES\_xl &1.5M &87.7 &67.7 &\underline{47.2} &\underline{80.3} &\underline{83.2} &\underline{43.3} &\underline{65.8} \\
        \hline
    \end{tabular}}
    \caption{Comparison of open-source math reasoning datasets with the FLAMES\_large and FLAMES\_xl datasets. DeepSeek-Math-7B is fine-tuned using unique problems from each dataset, alongside refreshed (R) solutions (using Qwen2.5-Math-7B-Instruct) for fair comparison. Resulting models are evaluated on in-domain, out-of-domain generalization (OOD), robustness, and competition benchmarks.}
    \label{tab:open-source-full}
\end{table*}

%% file: tables/Underlying_Models.tex
\begin{table*}[ht]
    \centering
    \setlength\tabcolsep{4pt} 
    \resizebox{\textwidth}{!}{ 
    \begin{tabular}{|c|c|c|c|c|c|c|c|c|}\hline
    & & \multicolumn{2}{|c|}{\textbf{In-Domain}} & \multicolumn{1}{|c|}{\textbf{OOD}} & \multicolumn{2}{c|}{\textbf{Robustness}} & \multicolumn{1}{c|}{\textbf{Competition}} & \\ \hline
    Student Model &Dataset &GSM8K &MATH5K &College Math &Distraction &GSMPlus &Olympiad Bench &Average \\ \hline
    DeepSeek-Math-7B &ScaleQuest Refreshed &\textbf{89.4} &66.0 &41.4 &71.0 &78.3 &25.0 &60.0 \\
    DeepSeek-Math-7B &FLAMES\_large &89.2 &\textbf{68.3} &41.9 &76.1 &79.4 &29.8 &\textbf{61.7} \\ \hline
    Qwen2.5-Math-7B &ScaleQuest Refreshed &\textbf{93.9} &80.7 &46.6 &78.4 &83.6 &41.3 &69.2 \\
    Qwen2.5-Math-7B &FLAMES\_large &93.5 &\textbf{81.4} &47.1 &82.9 &84.5 &40.9 &\textbf{69.5} \\ \hline
    Mathstral-7B &ScaleQuest Refreshed &88.4 &65.7 &40.0 &69.1 &76.5 &24.6 &59.0 \\
    Mathstral-7B &FLAMES\_large &\textbf{89.2} &\textbf{68.1} &39.9 &76.5 &79.3 &29.2 &\textbf{61.1} \\ \hline
    Mistral-7B-v0.3 &ScaleQuest Refreshed &84.7 &59.3 &36.7 &64.3 &73.5 &19.7 &54.8 \\
    Mistral-7B-v0.3 &FLAMES\_large &\textbf{86.4} &\textbf{63.6} &37.0 &73.8 &76.3 &24.1 &\textbf{57.5} \\ \hline
    Qwen2.5-14B &ScaleQuest Refreshed &93.1 &75.6 &44.6 &79.5 &83.5 &34.2 &66.2 \\
    Qwen2.5-14B &FLAMES\_large &\textbf{93.3} &\textbf{76.7} &44.9 &81.5 &84.0 &37.9 &\textbf{67.4} \\
    \hline
    \end{tabular}
    }
    \caption{Comparison of FLAMES\_large and ScaleQuest across diverse student models. DeepSeek-7B refers to DeepSeek-Math-7B, Mistral-7B refers to Mistral-7B-v0.3, Qwen2.5M-7B refers to Qwen2.5-Math-7B, FLAMES (L) refers to FLAMES\_large, ScaleQuest (R) refers to ScaleQuest problems with Qwen2.5-Math-7B-Instruct solutions. Both datasets contain 1M problems.}
    \label{tab:student}
\end{table*}

%% file: tables/DataMixture.tex
\begin{table*}[ht]
    \centering
    \setlength\tabcolsep{4pt} 
    \resizebox{\textwidth}{!}{ 
    \begin{tabular}{|c|c|c|c|c|c|c|c|c|c|}\hline
        & &\multicolumn{3}{|c|}{\textbf{In-Domain}} &\textbf{OOD} &\multicolumn{2}{|c|}{\textbf{Robustness}} &\textbf{Competition} & \\ \hline
        &\textbf{Agents} &\textbf{GSM8K} &\textbf{MATH5K} &\textbf{Average} &\textbf{College Math} &\textbf{Distraction} &\textbf{GSMPlus} &\textbf{Olympiad Bench} &\textbf{Average} \\ \hline
        &Distraction Insertion &83.3 &59.4 &71.3 &39.6 &\underline{72.4} &73.5 &24.7 &56.1 \\
        &IQC &\underline{86.0} &59.9 &73.0 &38.7 &72.1 &75.7 &24.4 &56.9 \\
        &Suggester-Editor &85.3 &\underline{61.0} &\underline{73.2} &39.4 &72.1 &\underline{75.9} &25.2 &57.4 \\
        &Taxonomy Key Concepts &77.1 &56.1 &66.6 &40.9 &64.7 &67.3 &21.8 &52.6 \\
        &QFT &79.0 &57.5 &68.3 &40.5 &67 &68.6 &23.6 &53.8 \\ \hline
        \textbf{Mixture ID} &\textbf{Agent Mixture} & & & & & & & &  \\ \hline
        A &\makecell{Suggester-Editor (50\%) \\ Taxonomy Key Concepts (50\%)} &82.9 &58.8 &70.9 &\underline{41.6} &69.0 &73.5 &23.7 &56.1 \\ \hline
        B &\makecell{Suggester-Editor (90\%) \\ Taxonomy Key Concepts (10\%)} &84.8 &60.0 &72.4 &41.1 &70.7 &74.8 &24.4 &57.0 \\ \hline
        C &\makecell{Suggester-Editor (75\%) \\ Taxonomy Key Concepts (25\%)}  &84.8 &59.4 &72.1 &41.4 &70.4 &74.6 &23.4 &56.7 \\ \hline
        D &\makecell{Suggester-Editor (50\%) \\ IQC (25\%) \\ Taxonomy Key Concepts (25\%)}  &84.9 &59.6 &72.3 &40.6 &69.5 &74.6 &25.2 &57.0 \\ \hline
        FLAMES Small &\makecell{Suggester-Editor (50\%) \\ Distraction Insertion (10\%) \\ IQC (20\%) \\ Taxonomy Key Concepts (20\%)} &85.2 &60.0 &72.6 &41.4 &72.2 &74.7 &\underline{26.1} &\underline{57.5} \\ \hline
        \end{tabular}
    }
    \caption{Results of underlying DeepSeek-Math-7B student model after fine-tuning on 150K problems of various agent data mixtures, alongside results using 150K from individual agents in each mixture.}
    \label{tab:mixture}
\end{table*}

%% file: tables/Solvability.tex
\begin{table}
    \centering
    \resizebox{0.45\textwidth}{!}{
        \begin{tabular}{|c|c|c|c|} \hline
        \textbf{Difficulty Level} &\textbf{Deemed Solvable} &\textbf{Total} &\textbf{\% Solvable} \\ \hline
        1 &37 &43 &86.00\% \\ \hline
        2 &75 &90 &83.30\% \\ \hline
        3 &81 &105 &77.10\% \\ \hline
        4 &88 &128 &68.80\% \\ \hline
        5 &68 &134 &50.70\% \\ \hline
        All &349 &500 &69.80\% \\ \hline
        \end{tabular}
    }
\caption{Number of real problems in the MATH500 evaluation dataset \cite{hendrycks2measuring} which were deemed unsolvable by the solvability filter proposed in \cite{scalequest}. We show in Section \ref{sec:quality-control-exp} that filtering synthetic problems for solvability does not lead to performance gains compared with simpler quality control measures.}
\label{tab:app:solvability}
\end{table}

%% file: tables/Appendix_Models.tex
\begin{table}
    \centering
    \resizebox{0.45\textwidth}{!}{
        \begin{tabular}{|c|c|c|} \hline
        \textbf{Model} &\textbf{Use Case}&\textbf{Relevant Sections} \\ \hline
        \makecell{Qwen2.5-32B-Instruct \\ \cite{yang2024qwen2}} &\makecell{FLAMES \\ Problem Generation} &Section \ref{sec:framework} \\ \hline
        \makecell{Qwen2.5-Math-7B-Instruct \\ \cite{yang2024qwen2}} &\makecell{FLAMES \\ Solution Generation} &Section \ref{sec:framework} \\ \hline
        \makecell{DeepSeek-Math-7B \\ \cite{shao2024deepseekmath}} &\makecell{FLAMES \\ Student Model} &Section \ref{sec:framework} \\ \hline
        \makecell{DeepSeek-v2.5 \\ \cite{deepseekai2024deepseekv2strongeconomicalefficient}} &\makecell{Ablation \\ Problem Generation} &Table \ref{tab:probsolmodels} \\ \hline
        \makecell{DeepSeek-Math-7B-RL \\ \cite{shao2024deepseekmath}} &\makecell{Ablation \\ Solution Generation} &Table \ref{tab:probsolmodels} \\ \hline
        \makecell{Qwen2.5-Math-7B \\ \cite{yang2024qwen2}} &\makecell{Ablation \\ Student Model} &Table \ref{tab:underlying} \\ \hline
        \makecell{Qwen2.5-14B \\ \cite{yang2024qwen2}} &\makecell{Ablation \\ Student Model} &Table \ref{tab:underlying} \\ \hline
        \makecell{Mathstral-7B \\ \cite{mistral_mathstral}} &\makecell{Ablation \\ Student Model} &Table \ref{tab:underlying} \\ \hline
        \makecell{Mistral-7B-v0.3\footnote{\url{https://huggingface.co/mistralai/Mistral-7B-v0.3}}} &\makecell{Ablation \\ Student Model} &Table \ref{tab:underlying} \\ \hline
        \end{tabular}
    }
\caption{Overview of all models used in the FLAMES framework, as well as models used in all controlled experiments (ablations).}
\label{tab:app:models}
\end{table}

%% file: tables/Appendix_Datasets.tex
\begin{table}
    \centering
    \resizebox{0.45\textwidth}{!}{
        \begin{tabular}{|c|c|c|} \hline
        \textbf{Dataset} &\textbf{Use Case}&\textbf{Relevant Sections} \\ \hline
        \makecell{GSM8K \\ \cite{cobbe2021training}} &\makecell{FLAMES Seed \\ Problems, Evalation} &Section \ref{sec:framework} \\ \hline
        \makecell{MATH \\ \cite{hendrycks2measuring}} &\makecell{FLAMES Seed \\ Problems, Evalation} &Section \ref{sec:framework} \\ \hline
        \makecell{GSMPlus \\ \cite{li2024gsm}} &\makecell{FLAMES Evaluation} &Section \ref{sec:framework} \\ \hline
        \makecell{CollegeMath\footnote{\url{https://huggingface.co/datasets/qq8933/College_Math_Test}}} &\makecell{FLAMES Evaluation} &Section \ref{sec:framework} \\ \hline
        \makecell{OlympiadBench\footnote{\url{https://huggingface.co/datasets/realtreetune/olympiadbench}}} &\makecell{FLAMES Evaluation} &Section \ref{sec:framework} \\ \hline
        \makecell{NuminaMath \\ \cite{numina_math_datasets}} &\makecell{Open-Source \\ Baseline} &Table \ref{tab:open-source} \\ \hline
        \makecell{MetaMathQA \\ \cite{metamath}} &\makecell{Open-Source \\ Baseline} &Table \ref{tab:open-source} \\ \hline
        \makecell{OrcaMath \\ \cite{orcamath}} &\makecell{Open-Source \\ Baseline} &Table \ref{tab:open-source} \\ \hline
        \makecell{OpenMathInstruct2 \\ \cite{omi2}} &\makecell{Open-Source \\ Baseline} &Table \ref{tab:open-source} \\ \hline
        \makecell{MMIQC \\ \cite{mmiqc}} &\makecell{Open-Source \\ Baseline} &Table \ref{tab:open-source} \\ \hline
        \makecell{ScaleQuest \\ \cite{scalequest}} &\makecell{Open-Source \\ Baseline} &Table \ref{tab:open-source} \\ \hline
        \end{tabular}
    }
\caption{Overview of all datasets used in the FLAMES framework.}
\label{tab:app:datasets}
\end{table}


%% file: tables/Appendix_Experiments.tex
\begin{table}
    \centering
    \resizebox{0.45\textwidth}{!}{
        \begin{tabular}{|c|c|} \hline
        \textbf{Experiment Description} &\textbf{Table(s)} \\ \hline
        \makecell{Comparison of Synthetic Data \\ Quality Control Strategies} &Table \ref{tab:quality-control} \\ \hline
        \makecell{Comparison of Existing \\ and Novel Data Agents} &Table \ref{tab:agents} \\ \hline
        \makecell{Comparison of FLAMES Datasets \\ with Existing Math Datasets} &\makecell{Table \ref{tab:open-source}, \\ Table \ref{tab:open-source-full}} \\ \hline
        \makecell{Comparison of FLAMES\_large \\ with Best Existing Dataset} &Table \ref{tab:underlying} \\ \hline
        \makecell{Comparison of Different Problem \\ and Solution Generation Models} &Table \ref{tab:probsolmodels} \\ \hline
        \makecell{Comparison of Different \\ Data Agent Mixtures} &Table \ref{tab:mixture} \\ \hline
        \makecell{Solvability Filtering \\ on MATH500 Test Set} &Table \ref{tab:app:solvability} \\ \hline
        \end{tabular}
    }
\caption{Overview of all controlled experiments used to design the FLAMES framework.}
\label{tab:app:experiments}
\end{table}

%% file: sections/AB_Prompt_Appendix.tex
\section{Prompts}
\label{app:agent-prompts}

In this section, we include prompts for agents evaluated in the FLAMES Framework (if not referenced in Table \ref{tab:app:agent-prompts}).

\begin{figure*}[t] 
\centering
\begin{fullwidthbox}
\textbf{Problem to Key Concepts Prompt:} You are a Maths teacher. I will give you a Maths problem and its solution. Your task is to tell a key concept which a student need to understand to solve this problem. We will use this key concept to create similar problems so that students can learn to solve similar problems. Your response should contain only the key concept.

Make sure you follow below constraints:

1. Your response is about generic concept and does not use specifics like numbers or words or object from the problem, so that the key concept can be used to generate related but different problems.

2. Your response provide granular details so that this response can be independently used to create a similar problem for teaching this key concept.

Below are few examples :

Problem : In how many ways can 5 students be selected from a group of 6 students?

Solution : We can choose 5 students out of a group of 6 students without regard to order in binom\{6\}\{5\} = 6 ways.

Key Concept : Number of ways to select some fixed number of items from a group of large number of items

Problem : Express frac\{3\}\{20\} as a decimal.

Solution : frac\{3\}\{20\} = 0.15.

Key Concept : Fraction to decimal conversion

Problem : A bicycle is traveling at 20 feet per minute. What is the bicycle's speed expressed in inches per second?

Solution : There are 12 inches in a foot, so the bicycle is traveling at 12(20)=240 inches per minute. There are 60 seconds in a minute, so the bicycle is traveling at frac\{240\}\{60\}=4 inches per second.

Key Concept : A math problem expressed as word problem which requires converting from one unit to another.

Problem : {problem}

Solution : {solution}

Key Concept : 
\end{fullwidthbox}
\caption{Prompt for extracting key concepts from existing GSM8K and MATH problems. Used in both Key Concepts and Seeded Key Concepts agents \cite{kpdds}.}
\end{figure*}

\begin{figure*}[t] 
\centering
\begin{fullwidthbox}
\textbf{Key Concepts Prompt:} You are a Maths teacher. I will give you an existing problem and a key concept which a student need to understand to solve this problem. Your task is to create a new Math problem that checks if a student understand this key concept and can use it to solve a Math problem. Do not include the solution in your new Math problem. Your response should contain only the new problem.

Make sure you follow below constraints:
1. The generated problem is grammatically correct, does not make any incorrect assumptions, is self-sufficient and can be solved without any additional information.

Below are few examples:

Key concept : Number of ways to select some items from a pool of large number of items.
New Problem : In how many ways can 4 students be selected from a group of 6 students?

Key Concept : Converting the units of a rate to new units
New Problem : A person is running a distance of 300 meters in 30 seconds. What is the speed of the person in feet per minute?

Key Concept : Using the triangle inequality to reason about triangles
New Problem : A stick 5 cm long, a stick 9 cm long, and a third stick $n$ cm long form a triangle. What is the sum of all possible whole number values of $n$?

Key Concept : {keyconcept}
New Problem : 
\end{fullwidthbox}
\caption{Prompt for using key concepts to generate synthetic problem. Used by Key Concepts agent \cite{kpdds}.}
\end{figure*}

\begin{figure*}[t] 
\centering
\begin{fullwidthbox}
\textbf{Seeded Key Concepts Prompt:} You are a Maths teacher. I will give you an existing problem and a key concept which a student need to understand to solve this problem. Your task is to create a new Math problem that checks if a student understand this key concept and can use it to solve a Math problem. Do not include the solution in your new Math problem. Your response should contain only the new problem.

Make sure you follow below constraints:
1. The generated problem is grammatically correct, does not make any incorrect assumptions, is self-sufficient and can be solved without any additional information. 
2. The generated problem is of similar complexity as the existing problem. 

Below are few examples:

Existing Problem : Determine the number of ways to arrange the letters of the word ELLIPSE.
Key concept : Number of ways to select some items from a pool of large number of items.
New Problem : In how many ways can 4 students be selected from a group of 6 students?

Existing Problem : A bicycle is traveling at 20 feet per minute. What is the bicycle's speed expressed in inches per second?
Key Concept : Converting the units of a rate to new units
New Problem : A person is running a distance of 300 meters in 30 seconds. What is the speed of the person in feet per minute?

Existing Problem : The lengths of the sides of a non-degenerate triangle are x, 13 and 37 units. How many integer values of x are possible?
Key Concept : Using the triangle inequality to reason about triangles
New Problem : A stick 5 cm long, a stick 9 cm long, and a third stick n cm long form a triangle. What is the sum of all possible whole number values of n?

Existing Problem : {problem}
Key Concept : {keyconcept}
New Problem : 
\end{fullwidthbox}
\caption{Prompt for using problem and key concepts to generate synthetic problem. Used by Seeded Key Concepts agent \cite{kpdds}.}
\end{figure*}

\begin{figure*}[t] 
\centering
\begin{fullwidthbox}

\textbf{Distraction Insertion Prompt:} Your goal is to hide a misleading detail in the given problem, such that it doesn't change the answer to the problem. The solution to the new problem should be the same as the solution to the original problem. Do not give a solution for the new problem. Do not give solution hints.

Here is an example:

Problem: Natalia sold clips to 48 of her friends in April, and then she sold half as many clips in May. How many clips did Natalia sell altogether in April and May?

Solution: Natalia sold altogether 72 clips in April and May.

New Problem: Natalia sold clips to 48 of her friends in April, and then she sold half as many clips in May. In June, she sold twice as many clips as in April. How many clips did Natalia sell altogether in April and May?

Now hide a misleading detail in the following problem.

Problem: {problem}

Solution: {solution}

New Problem:
\end{fullwidthbox}
\caption{Prompt used to generate synthetic problems using the novel Distraction Insertion agent (see Section \ref{sec:data-agents}).}
\end{figure*}

\begin{figure*}[t] 
\centering
\begin{fullwidthbox}

\textbf{Solvability Filtering Prompt:} Please act as a professional math teacher.
Your goal is to determine if the given problem is a valuable math problem. You need to consider two
aspects:

1. The given problem is a math problem.

2. The given math problem can be solved based on the conditions provided in the problem (You can first
try to solve it and then judge its solvability).

Please reason step by step and conclude with either ‘Yes’ or ‘No’.

Given Problem: \{problem\}
\end{fullwidthbox}
\caption{Prompt used to filter unsolvable synthetic problems \cite{scalequest}, see Section \ref{sec:quality-control}.}
\end{figure*}